\documentclass[sigconf]{acmart}

\usepackage{amsmath}
\usepackage{algorithm}
\usepackage{algorithmic}
\usepackage{threeparttable} 
\definecolor{baselinecolor}{gray}{.9}
\definecolor{deemph}{gray}{0.6}

\newcommand{\gc}[1]{\textcolor{deemph}{#1}}
\usepackage{multirow}
\usepackage{subfigure}
\usepackage{colortbl}  
\usepackage{amsmath}
\usepackage[normalem]{ulem}
\usepackage[bottom]{footmisc}
\usepackage{footmisc}

\AtBeginDocument{%
  \providecommand\BibTeX{{%
    \normalfont B\kern-0.5em{\scshape i\kern-0.25em b}\kern-0.8em\TeX}}}

\copyrightyear{2023} 
\acmYear{2023} 
\setcopyright{acmlicensed}\acmConference[KDD '23]{Proceedings of the 29th ACM SIGKDD Conference on Knowledge Discovery and Data Mining}{August 6--10, 2023}{Long Beach, CA, USA}
\acmBooktitle{Proceedings of the 29th ACM SIGKDD Conference on Knowledge Discovery and Data Mining (KDD '23), August 6--10, 2023, Long Beach, CA, USA}
\acmPrice{15.00}
\acmDOI{10.1145/3580305.3599445}
\acmISBN{979-8-4007-0103-0/23/08}

\begin{document}

\title{Neural-Hidden-CRF: A Robust Weakly-Supervised \\
Sequence Labeler}

\author{Zhijun Chen}
\affiliation{%
  \institution{SKLSDE Lab, Beihang University}
  \city{Beijing}
  \country{China}}
\email{zhijunchen@buaa.edu.cn}

\author{Hailong Sun}
\authornote{Corresponding author.}
\affiliation{%
  \institution{SKLSDE Lab, Beihang University}
   \city{Beijing}
  \country{China}}
\email{sunhl@buaa.edu.cn}

\author{Wanhao Zhang}
\affiliation{%
  \institution{Institute for Network Sciences and Cyberspace, Tsinghua University}
   \city{Beijing}
  \country{China}}
\email{zwh22@mails.tsinghua.edu.cn}

\author{Chunyi Xu}
\affiliation{%
  \institution{SKLSDE Lab, Beihang University}
   \city{Beijing}
  \country{China}}
\email{zy2106118@buaa.edu.cn}

\author{Qianren Mao}
\affiliation{%
 \institution{Zhongguancun  Laboratory}
   \city{Beijing}
  \country{China}}
\email{maoqr@zgclab.edu.cn}

\author{Pengpeng Chen}
\affiliation{%
  \institution{China’s Aviation System Engineering Research Institute}
   \city{Beijing}
  \country{China}}
\email{chenpp@buaa.edu.cn}

\renewcommand{\shortauthors}{Zhijun Chen, et al.}


\begin{abstract}
We propose a neuralized undirected graphical model called Neural-Hidden-CRF to solve the weakly-supervised sequence labeling problem.
Under the umbrella of probabilistic undirected graph theory, the proposed Neural-Hidden-CRF embedded with a hidden CRF layer models the variables of word sequence, latent ground truth sequence, and weak label sequence with the global perspective that undirected graphical models particularly enjoy.
In Neural-Hidden-CRF, we can capitalize on the powerful language model BERT or other deep models to provide rich contextual semantic knowledge to the latent ground truth sequence, and use the hidden CRF layer to capture the internal label dependencies.
Neural-Hidden-CRF is conceptually simple and empirically
powerful.
It obtains new state-of-the-art results on one crowdsourcing benchmark and three weak-supervision benchmarks, including 
outperforming the recent advanced model CHMM by 2.80 F1 points and 2.23 F1 points in average generalization and inference performance, respectively.
\end{abstract}



\maketitle

\section{Introduction}
\label{introduction}

Deep learning has witnessed the insatiable appetite for humongous labeled training data.
This appetite for data motivated several lines of work, such as active learning~\cite{matsushita2018deep}, semi-supervised learning~\cite{ouali2020overview},  transfer learning~\cite{weiss2016survey}, and more recently, \textit{weak supervision} (WS) ~\cite{zhang2021wrench, zhang2022survey}, which is of interest in this paper.

As a time/cost-efficient and easy-to-promote alternative to gold expert annotation, WS provides practitioners with \textit{multiple heterogeneous weak supervision sources}, such as crowdsourcing annotators from the Internet, user-defined programs encoded external knowledge bases, patterns/rules, or pre-trained classifiers, etc~\cite{ratner2017snorkel,zhang2021wrench, zhang2022survey}.
As the price of good accessibility and as the name ``weak supervision'' implies, these various weak sources often exhibit varying error rates, leading to the generation of conflicting and noisy labels in many instances.

WS has been applied to various tasks, including the fundamental deep language understanding task---\textit{sequence labeling}~\cite{ma2016end}, whose importance has been well recognized in the natural language processing
community.
In this paper, we focus on the problem of sequence learning in the context of multiple heterogeneous weak supervision sources, which can be abbreviated as \textit{weakly-supervised sequence labeling} (WSSL).
It has been extensively studied as another main research branch in the whole WS community in addition to normal independent classification tasks~\cite{zhang2021wrench, zhang2022survey}, because of the importance of the sequence labeling problem itself and the challenges associated with the need to consider the internal dependencies among sequence labels when solving WSSL.

To address the WSSL problem, existing representative methods fall into three categories in intrinsic methodology:
\begin{itemize}
\item
The HMM-based graphical models~\cite{nguyen2017aggregating,simpson2018bayesian, safranchik2020weakly, lison2020named, lison2021skweak} leverage the hidden Markov model (HMM)~\cite{blunsom2004hidden} to model the generation process of latent truth label sequence and observed weak label sequence, and then apply the expectation maximization (EM) algorithm~\cite{moon1996expectation} to infer truth labels.
(Then, these inferred labels, in turn, can be used to train a final sequence labeler.)
Though principled, these models fall short in leveraging token semantics and context information~\cite{li2021bertifying}, as they either model input tokens as one-hot observations~\cite{nguyen2017aggregating,simpson2018bayesian} or do not model them at all~\cite{safranchik2020weakly, lison2020named, lison2021skweak}.
\item 
The ``source-specific perturbation'' deep learning models~\cite{nguyen2017aggregating, lan2019learning, zhang2021crowdsourcing}, train multiple weak source-specific deep models, obtained by inserting the \textit{source-specific perturbation parameters} to the \textit{unique shared deep model parameters}, and perform test using the \textit{assumed optimal classifier} obtained by the shared deep model straightforwardly or a certain combination of the source-specific deep models.
With less principle, it is not clear how interpretable they are in terms of mechanism design.
\item 
The recently proposed neuralized HMM-based graphical models~\cite{li2021bertifying, li2022sparse} construct HMM-based directed graphical models in which the dependencies among variables of word sequence, latent ground truth  sequences and weak labels are sophisticatedly modeled, and rich contextual semantic information is introduced using deep learning techniques (e.g., the language model BERT~\cite{devlin2018bert}).
\end{itemize}

The neural HMM-based graphical models have the methodological advantages of both the first two classes of approaches---i.e., the principled modeling of graphical models to model variable dependencies and the rich contextual knowledge that comes from using deep learning---and have achieved relatively most satisfactory performance empirically in the recent WS benchmark~\cite{zhang2021wrench}.
However, these methods internally split all the variables of interest into multiple \textit{local} regions and model them separately, and separately model the conditional probabilities of the ground truth at each time step ($p(t_l \mid t_{l-1}, \mathbf{x})$) in the truth sequence. 
Essentially, this \textit{per-state normalization} approach~\cite{lafferty2001conditional} (coming from the per-step modeling) is the same as that of the MEMM model~\cite{mccallum2000maximum}, directly making them suffer from the well-known thorny \textit{label bias problem}~\cite{lafferty2001conditional} that is often mentioned in sequence labeling problems~\cite{sutton2012introduction, simoes2009information, wallach2004conditional}.
In short, this approach of using the local optimization perspective (coming from repeatedly considering patterns for the scale of a step instead of holistically considering the entire sequence) causes some useful information to be erased~\cite{hannun2020label} and leads to some bias.
In fact, mainly because of this reason, the  canonical conditional random field (CRF)~\cite{lafferty2001conditional} was deliberately proposed by scholars in order to solve the sequence labeling problem with a globalized perspective.

In this paper, we move one step further and explore: \textit{when solving the WSSL problem, how can we capitalize on the graphical model with principled modeling of variable dependencies and the advanced deep learning model that can bring rich contextual knowledge, without introducing the label bias problem, in a unified model?}
To address this problem, we introduce Neural-Hidden-CRF, a neuralized graphical model embedded with a hidden CRF layer.
Neural-Hidden-CRF is built on undirected graph theory and models three sets of variables---namely, word sequence, latent ground truth sequence, and weak label sequence---with a globalized perspective like CRFs instead of the HMM-based models always considering local knowledge.
Specifically, in Neural-Hidden-CRF, we use deep learning models (like the language model BERT) to flexibly transfer rich contextual semantic knowledge to the latent truth sequence, and use the embedded hidden CRF layer to capture the dependencies among the truth sequences, and use the \textit{weak source transition matrices} to model the dependencies between the truth labels and the weak labels.
By doing so, our model benefits both from the expressiveness and reasonableness of graphical models for capturing sophisticated dependencies among variables and from the effectiveness of the deep learning models for obtaining  contextual semantic knowledge, while avoiding the label bias problem caused by the local  perspective.
To the best of our knowledge, this is the first work to apply a neuralized undirected graphical model to solve the WSSL problem.
We conduct extensive evaluations of the proposed Neural-Hidden-CRF on one crowdsourcing benchmark and three WS benchmarks, showing that Neural-Hidden-CRF is a robust weakly-supervised sequence labeler and outperforms the state-of-the-art.
\footnotemark
\footnotetext{The code is available at: 
\url{https://github.com/junchenzhi/Neural-Hidden-CRF}.}

\subsection{Related Work}
\textbf{WSSL Learning Paradigms.}
To address the WSSL problem, two learning paradigms exist~\cite{zhang2021wrench}: 
(1) Two-stage paradigm: Researchers have developed \textit{label models}~\cite{nguyen2017aggregating,simpson2018bayesian, safranchik2020weakly, lison2020named, lison2021skweak}  (also known as \textit{truth inference} models~\cite{zheng2017truth}) to aggregate noisy weak labels for each instance, accomplished with an follow-up \textit{end model} (i.e., classifier) learning process using the aggregated labels;
(2) Joint paradigm: Later researchers also explored learning the classifier of interest directly from weak supervision labels through ad hoc \textit{joint models} in an end-to-end manner. 
As presented above, we categorize representative methods from the intrinsic methodological perspective, where each method mentioned in their original work is either emphasized for its truth inference capability~\cite{simpson2018bayesian, safranchik2020weakly, lison2020named, lison2021skweak} or generalization performance~\cite{lan2019learning} or, better yet, both~\cite{nguyen2017aggregating, zhang2021crowdsourcing}.

\textbf{Other WSSL Works.}
All WSSL methods can be divided into probabilistic graphical model approach, deep learning model approach, and neuralized graphical model approach.
(1) In probabilistic graphical model approach (and in addition to the HMM-based models~\cite{nguyen2017aggregating,simpson2018bayesian, safranchik2020weakly, lison2020named, lison2021skweak}), \citet{rodrigues2014sequence} in early 2014 used a partially directed graph containing a CRF for modeling to solve the truth inference from crowdsourcing labels;
(2) In deep learning model approach (and in addition to the ``source-specific perturbation'' methods~\cite{nguyen2017aggregating, lan2019learning, zhang2021crowdsourcing}), other methods~\cite{rodrigues2018deep, sabetpour2020optsla, sabetpour2021truth, lan2019learning} are either based on the end-to-end deep neural architecture~\cite{rodrigues2018deep}, or the customized optimization objective along with coordinate ascent optimization technology~\cite{sabetpour2020optsla, sabetpour2021truth}, or the iterative solving framework similar to expectation–maximization algorithm~\cite{chen2023learning}.
However, all these methods do not have the advantages of the recently proposed neuralized HMM-based graphical models~\cite{li2021bertifying, li2022sparse} and our Neural-Hidden-CRF in principled modeling for variants of interest and in harnessing the context information that provided by advanced deep learning models.
Additionally, it is worth mentioning the presence of numerous established WS methods that address the normal independent classification scenario~\cite{zhang2021wrench,zhang2022survey,zhang2022knowledge,chen2020structured,chen2022adversarial}.

\begin{figure*}[ht]
\centering
\includegraphics[width=0.49\textwidth]{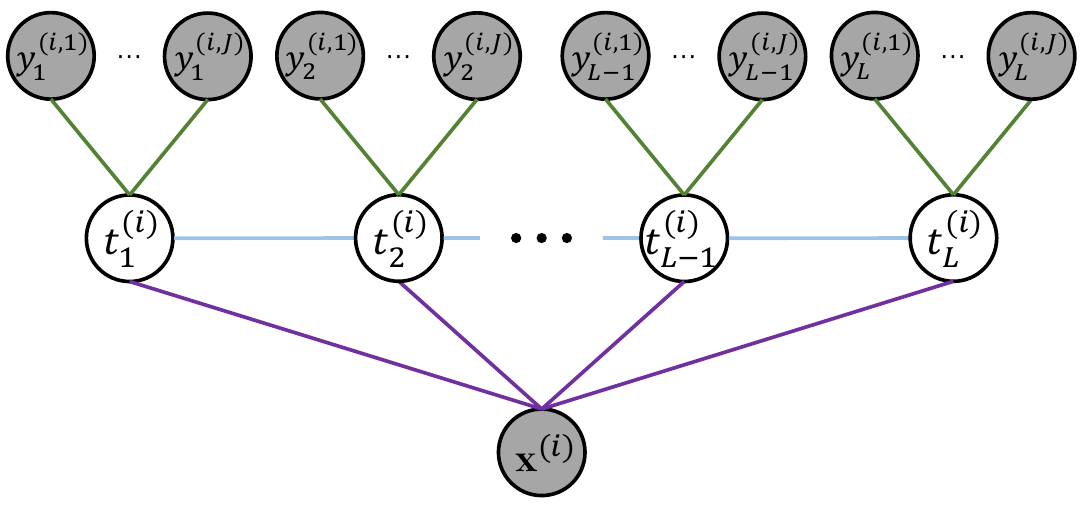} 
\caption{Probabilistic graphical representation of Neural-Hidden-CRF.}
\label{Figure:1}
\end{figure*}

\section{Neural-Hidden-CRF}

\subsection{Problem Formulation and Preliminaries}
\label{Problem Formulation and Preliminaries}

\paragraph{Problem Formulation of WSSL}
We are given i.i.d. training data $\mathcal{D}=\{\mathbf{x}^{(i)}, \mathbf{y}^{(i)}\}_{i=1}^{I}$, where $\mathbf{x}^{(i)}=\{x_{l}^{(i)}\}_{l=1 }^{L} \in \mathcal{X}$ is an observed word sequence with $L$-length tokens,  $\mathbf{y}^{(i)}=\{y_{l}^{(i, j)}\}_{j \in \mathcal{J}^{(i)}, l  \in {(1,2, \ldots, L)}}$ are the noisy weak labels attached to  $i^{\text {th }}$ sentence, and $\mathcal{J}^{(i)}$ represents the set of weak sources that labeled the $i^{t h}$ sentence among all $J$ sources.
For each sentence $\mathbf{x}^{(i)}$, there is a latent ground truth sequence $\mathbf{t}^{(i)}=\{t_{l}^{(i)}\}_{l=1 }^{L}\in\mathcal{T}$ \textit{unobserved} to us; $t_{l}^{(i)}\in\{1,2, \ldots, K\}$, where $K$ denotes  number of categories.
(In addition,  we use $y^{(i, j)}_l=0$ to denote that source $j^{t h}$ has not annotated sentence $\mathbf{x}^{(i)}$.)
Our goal is to learn from the weak supervision data $\mathcal{D}=\{\mathbf{x}^{(i)}, \mathbf{y}^{(i)}\}_{i=1}^{I}$ to obtain a sequence labeler $f: \mathcal{X} \mapsto \mathcal{T}$ with strong generalization.

\paragraph{Preliminaries on Undirected Graphical Models}
Here we give the most fundamental overview of the underlying theory (as an optional reading part).
\textbf{(1)}
The set of nodes in an undirected graph, 
where edges connect any two nodes, is denoted as a \textit{cluster}; if a cluster cannot be added to any node to make it a larger cluster, it is denoted as a \textit{maximum cluster}~\cite{klinger2007classical}.
\textbf{(2)}
Under the undirected graphical model theory, the probability distribution over all nodes is factored as the normalized product of \textit{potential functions} of all maximum clusters~\cite{klinger2007classical}:
\begin{equation}
\begin{array}{c}
p(X)=\frac{1}{Z} \prod_c \phi_c\left(X_c\right) \\
Z=\sum_X \prod_c \phi_c\left(X_c\right),
\end{array}
\end{equation}
where $X_c$ is the set of nodes in a maximum cluster $c$, $\phi_c\left(\cdot\right)$ is an arbitrary non-negative real-valued function, called the \textit{potential function} (acting as a scoring role), and $Z$ is the normalization factor.

\subsection{Model}
We first formally introduce our model Neural-Hidden-CRF in Sections~\ref{Model}-~\ref{Eventually Neuralized Model}, and briefly explain Neural-Hidden-CRF from another simpler vision in Section~\ref{Understanding Our Model from a Simpler Perspective}.
The graphical representation of  Neural-Hidden-CRF is shown in Figure~\ref{Figure:1}. 
Note that in Figure~\ref{Figure:1} and the presentation (except for the derivation in the Appendix~\ref{Calculation of the Likelihood}) that follows, we tacitly assume that each instance owns annotations from all weak supervision sources.

\subsubsection{Model}
\label{Model}
In order to present in a more understandable way, we first introduce the base version of the model Neural-Hidden-CRF and then introduce the process of its neuralization in Section~\ref{Eventually Neuralized Model} to obtain the eventual neuralized version, i.e., the model we refer to by default.
Previously, the CRF~\cite{lafferty2001conditional} was upgraded to its neuralized version (e.g., BiLSTM-CRF~\cite{huang2015bidirectional} or BERT-CRF~\cite{zhang2021wrench}) by the same neuralization process.

Similar to the original CRF theory \cite{lafferty2001conditional} and corresponding to Figure~\ref{Figure:1}, we define the probability of weak label sequence $\mathbf{y}^{(i)}$ along with   ground truth sequence  $\mathbf{t}^{(i)}$ given observation sequence  $\mathbf{x}^{(i)}$ to be a normalized product of three kinds of \textit{pseudo-potential function}\footnote{Note that for the sake of clarity, here we do not start with the construction of our \textit{potential functions} in the strict sense, as mentioned in Section~\ref{Problem Formulation and Preliminaries}.
Instead, we create and define the ``pseudo-potential function'', which is similar to the strict \textit{potential function}, but its role is not equivalent to that of a strictly defined one.} (which can undergo some simple adaptions to form the strictly potential functions, as mentioned in Seciton~\ref{Problem Formulation and Preliminaries}), i.e.,
\begin{equation}
\exp \left(\sum_{a} \lambda_{a}\text{state}_{a}(t^{(i)}_{l}, \mathbf{x}^{(i)}, l)\right),
\label{equation 2}
\end{equation}

\begin{equation}
\exp \left(\sum_{b} \mu_{b}\text{transition}_{b}(t^{(i)}_{l-1}, t^{(i)}_{l}, l)\right),
\label{equation 3}
\end{equation}

\begin{equation}
\exp \left(\sum_{c,j} \eta_{c,j}\text{source}_{c,j}(y^{(i,j)}_l, t^{(i)}_l, l)\right),
\label{equation 4}
\end{equation}
where $l\in {(1,2, \ldots, L)}$ denotes the time step.
For now, it is sufficient to note that: 
\textbf{(1)} Within these \textit{pseudo-potential functions}, each of the \textit{feature functions} $\text{state}_{a}(\cdot)$, $\text{transition}_{b}(\cdot)$, $\text{source}_{c,j}(\cdot)$---aiming at extracting features---can be the pre-defined indicator function that takes the value 1 when the internal declaration is satisfied, and 0 otherwise; the corresponding \textit{weights} $\lambda_{a}, \mu_{b}, \eta_{c,j}\in(-\infty,+\infty)$ are model's parameters to be estimated;
\textbf{(2)} These exponential \textit{pseudo-potential functions} play the role of scoring (for a specific instantiated value of $\mathbf{x}^{(i)}$, $\mathbf{t}^{(i)}$, $\mathbf{y}^{(i)}$ on every time step) and will be non-negative.
In other words, the internal \textit{feature functions} along with their  weights play the role of scoring;
\textbf{(3)} Intuitively, the three types of \textit{feature functions},  $\text{state}_{a}(\cdot)$, $\text{transition}_{b}(\cdot)$, $\text{source}_{c,j}(\cdot)$---corresponding to  purple lines,  blue lines, and  green lines in  Figure~\ref{Figure:1}---act between the token sequence  $\mathbf{x}^{(i)}$ and the  truth $\mathbf{t}^{(i)}$, between the interior of the truth sequence $\mathbf{t}^{(i)}$, and between the truth  $\mathbf{t}^{(i)}$ and the weak labels
$\mathbf{y}^{(i)}$, respectively.
Specifically, we will walk through more details about \textit{features functions} in Section~\ref{Feature Functions}.

For simplicity, we denote the above three \textit{feature functions} and the corresponding  \textit{weights} by the general notations:
\begin{equation}
\begin{aligned}
&f_{w}(\mathbf{y}^{(i)}_{l},t^{(i)}_{l-1},t^{(i)}_{l},\mathbf{x}^{(i)}, l)=\\
&\left\{\begin{array}{l}
\text{state}_{a}(t^{(i)}_{l}, \mathbf{x}^{(i)}, l) \quad \qquad  \qquad
\text{if } w=1,.., A\\
\text{transition}_{b}(t^{(i)}_{l-1}, t^{(i)}_{l}, l) \qquad 
\enspace
\;   \, \text{if } w=A+1,.., A+B \\
\text{source}_{c,j}(y^{(i,j)}_l, t^{(i)}_l, l)   \quad \qquad \text{ if } w=A+B+1,.., A+B+\sum_j C_j,
\end{array}\right.
\end{aligned}
\label{equation 5}
\end{equation}
along with
\begin{equation}\theta_{w}=\left\{\begin{array}{ll}\lambda_{a} & \text{if } w=1,.., A \\\mu_{b} & \text{if } w=A+1,..., A+B \\
\eta_{c,j} & \text{if } w=A+B+1,..., A+B+\sum_j C_j,
\end{array}\right.
\label{equation 6}
\end{equation}
where $A$, $B$, and $\sum_j C_j$ denote the specific number of a certain type of feature function, respectively.

Thus, our conditional model can be expressed as:
\begin{equation}
\begin{aligned}
&  p(\mathbf{y}^{(i)}, \mathbf{t}^{(i)} \mid \mathbf{x}^{(i)}; \Theta) \\
=&\frac{1}{\boldsymbol{Z}(
\mathbf{x}^{(i)}; \Theta)} \exp \left(\sum_{l}\sum_{w} \theta_{w} \cdot f_{w}(\mathbf{y}^{(i)}_{l},
t^{(i)}_{l-1},
t^{(i)}_{l},
\mathbf{x}^{(i)}, l)\right),
\end{aligned}
\label{equation 7}
\end{equation}
where $\boldsymbol{Z}(\mathbf{x}^{(i)}; \Theta)$ is the instance-specific normalization factor (also called \textit{partition function} in the CRF~\cite{lafferty2001conditional}) defined as:
\begin{equation}
\boldsymbol{Z}(\mathbf{x}^{(i)}; \Theta)=\sum_{\mathbf{y}^{(i)}} \sum_{\mathbf{t}^{(i)}} \exp \left(\sum_{l}\sum_{w} \theta_{w} \cdot f_{w}(\mathbf{y}^{(i)}_{l}, t^{(i)}_{l-1}, t^{(i)}_{l}, \mathbf{x}^{(i)}, l)\right).
\label{equation 8}
\end{equation}
More intuitively, if  $\exp \left(\sum_l \sum_w \theta_w \cdot f_w(\mathbf{y}_l^{(i)}, t_{l-1}^{(i)}, t_l^{(i)}, \mathbf{x}^{(i)}, l)\right)$, representing getting the score for a specific instantiation of ($\mathbf{y}^{(i)}$, $\mathbf{t}^{(i)}$, $\mathbf{x}^{(i)}$), abbreviated with 
$\Phi(\mathbf{y}^{(i)},\mathbf{t}^{(i)},\mathbf{x}^{(i)})$, then Equation~\ref{equation 7} can be rewritten as:
\begin{equation}
p(\mathbf{y}^{(i)}, \mathbf{t}^{(i)} \mid \mathbf{x}^{(i)}; \Theta) = 
\frac{\Phi(\mathbf{y}^{(i)}, \mathbf{t}^{(i)}, \mathbf{x}^{(i)})}{\sum_{\mathbf{y}^{(i)}} \sum_{\mathbf{t}^{(i)}} \Phi\left(\mathbf{y}^{(i)}, \mathbf{t}^{(i)}, \mathbf{x}^{(i)}\right)}.
\label{equation 9}
\end{equation}
Essentially, our model given by Equation~\ref{equation 7} or Equation~\ref{equation 9} is inherently aligns with the underlying theory mentioned in Section~\ref{Problem Formulation and Preliminaries}.\footnote{Referring to more material on CRFs and our tutorial
(\url{https://github.com/junchenzhi/Neural-Hidden-CRF}) would help to enhance the comprehension of our model.}

\paragraph{The Embodiment of the Global Optimization Perspective}
We can notice that the our pseudo-potential functions and feature functions in Equations~\ref{equation 2}-\ref{equation 3} do not have a direct probabilistic interpretation, but instead represent constraints or scores on the configurations of the random variable.
As a result, the model expressed by Equation~\ref{equation 7} yields a \textit{global normalized score} for $p(\mathbf{y}^{(i)}, \mathbf{t}^{(i)} \mid \mathbf{x}^{(i)} ; \Theta)$.
This \textit{global normalization} approach is unlike all HMMs, where $(\mathbf{y}^{(i)}, \mathbf{t}^{(i)})$ is split into multiple uni-directional dependent random variables (i.e., a set consisting of many $(y^{(i,j)}_l, t^{(i)}_l)$) based on some strict independence assumptions and each conditional probability distribution between random variables is normalized (e.g., \textit{local normalization} for per-step in~\citet{li2021bertifying, li2022sparse}) to further obtain the probability of the joint distribution $(\mathbf{y}^{(i)}, \mathbf{t}^{(i)})$.
Simply put, our approach models/trains holistically (the learned knowledge is global), while the HMMs~\cite{li2021bertifying, li2022sparse} decompose the modeling into multiple uni-directional dependent local regions and model the patterns for the scale of a step (the learned knowledge is local).
As a result of the holistic undirected graphical modeling, our method can result in model parameters that are not constrained by probabilistic forms, thus enjoying more flexible scoring.
As for the \textit{label bias problem}, our model circumvents this  by adopting the \textit{global normalization} rather than the \textit{local normalization} in~\citet{li2021bertifying, li2022sparse}, just as the CRF model does with respect to the MEMM model~\cite{lafferty2001conditional,hannun2020label}.
Please refer to~\citet{hannun2020label} for more information on the label bias problem.

\subsubsection{Feature Functions}
\label{Feature Functions}
(1) For feature function $\text{state}_{a}(\cdot)$, like the original CRF theory, we can define the following example:
\begin{small}
\begin{equation}
\text{state}_{a}(t^{(i)}_{l}, \mathbf{x}^{(i)}, l)=\left\{\begin{array}{ll}1 & \text { if } t^{(i)}_{l}= \texttt{PERSON}
\text { and } x^{(i)}_{l}=\texttt{John} \\0 & \text { otherwise}.
\end{array}\right.
\label{equation 10}
\end{equation} 
\end{small}
If the corresponding weight $\lambda_a$ is a relatively large value, whenever the internal declaration in $\text{state}_a(\cdot)$ is true, it increases the probability of the sequence $\mathbf{t}^{(i)}$.
Intuitively, the model would prefer the tag \texttt{PERSON} for the word \texttt{John}.
Formally, whenever the internal declaration in $\text{state}_{a}(\cdot)$ is satisfied, this feature function along with its weights $\lambda_{a}$ will contribute factor $\exp (\lambda_{a} \cdot 1)$ to the numerator in Equation~\ref{equation 7}.

(2) For feature function $\text{transition}_{b}(\cdot)$, we can define the following example:
\begin{small}
\begin{equation}
\text{transition}
_{b}(t^{(i)}_{l-1}, t^{(i)}_{l}, l)
=\left\{\begin{array}{ll}1 & \text {if } 
t^{(i)}_{l-1}=\texttt{OTHER} \text{ and }
t^{(i)}_{l}=\texttt{PERSON} \\
0 & \text {otherwise}.
\end{array}\right.
\end{equation}
\label{equation 11}
\end{small}
Also, whenever the internal declaration in $\text{transition}_{b}(\cdot)$ is satisfied, this feature function along with its weights $\mu_{b}$ will contribute factor $\exp (\mu_{b} \cdot 1)$ to the numerator in Equation~\ref{equation 7}.
Since the number of categories is $K$, naturally we can define all $K^{2}$ feature functions of $\text{transition}_{b}(\cdot)$.
The set of parameters concerning $\text{transition}_{b}(\cdot)$ can referred to as the \textit{CRF transition matrix} (with size of $K \times K$), which essentially captures the dependence within the label sequence.

(3) For feature function $\text{source}_{c,j}(\cdot)$, we can define the following example:
\begin{small}
\begin{equation}
\text{source}
_{c,j}(y^{(i,j)}_{l}, t^{(i)}_{l}, l)
=\left\{\begin{array}{ll}
1 & \text { if } 
y^{(i,j)}_{l}=\text {PERSON and }
t^{(i)}_{l}=\text {PERSON} \\
0 & \text { otherwise}.
\end{array}\right.
\end{equation}
\label{equation 12}
\end{small}
Similar to feature function $\text{transition}_{b}(\cdot)$, we can define all $K^{2}$ feature functions of $\text{source}_{c,j}(\cdot)$ for weak source $j^{th}$.
For the particular $j^{\text {th}}$ source, weights $\{\eta_{c,j}\}_{c=1 }^{K^{2}}$ naturally form a $K \times K$ matrix that can represents the behavior pattern of this source.
Thus, the higher the ability of a source, the larger the value of the diagonal elements of its matrix relative to the value of the non-diagonal elements.
Similar to the \textit{CRF transition matrix}, we can refer to  this matrix as \textit{weak source transition matrix}.

\subsubsection{Eventually Neuralized Model}
\label{Eventually Neuralized Model}
Here we introduce a deep sequence  network, such as the language model BERT without the last softmax layer, between sequence $\mathbf{x}$ and sequence $\mathbf{t}$ to complete the model's neuralization.
Thus: (\textit{i}) In the basic version of the model described above, for the time step $l$ in the sentence $\mathbf{x}^{(i)}$, the feature function $\text{state}_{a}(\cdot)$ along with it's weight $\lambda_{a}$ provide the factor $\exp (\lambda_{a} \cdot \text { state}_{a}(t_{l}^{(i)}, \mathbf{x}^{(i)}, l)) $---which represents the degree of support of the model for ($\mathbf{x}^{(i)}_{l}, t^{(i)}_{l} $)---for the numerator in Equation~\ref{equation 7};
(\textit{ii}) In the current neuralized model, we use
\begin{equation}
\text {extract}_{l, t^{(i)}_{l}}(\sigma_{BERT}(\mathbf{x}^{(i)})),
\label{equation 13}
\end{equation}
which also represents the degree of support of the model for ($\mathbf{x}^{(i)}_{l}, t^{(i)}_{l} $), as the factor provided to the numerator in Equation~\ref{equation 7}.
In Equation~\ref{equation 13}, $\sigma_{BERT}(\cdot)$ is the output logits of BERT, 
and $\operatorname{extract}_{l, t_{l}^{(i)}}(\cdot)$ extracts the probability mass of the category $t^{(i)}_{l}$ in the $l$-step element of the input.

\subsubsection{Understanding Our Model from a Simpler Perspective.}
\label{Understanding Our Model from a Simpler Perspective}

Our Neural-Hidden-CRF, for weakly-supervised sequence labeling learning, shares similarities with CRFs (e.g., BERT-CRF), for supervised sequence labeling learning.
We show the graphical representation of CRF vs. Neural-Hidden-CRF in Appendix~\ref{Probabilistic Graphical Representation}.
\textbf{(1)}
First, BERT-CRF is a discriminative model concerning the  label sequence $\mathbf{t}^{(i)}$  given the sentence $\mathbf{x}^{(i)}$:
\begin{equation}
p(\mathbf{t}^{(i)} \mid \mathbf{x}^{(i)}; \Theta)=\frac{\exp (\operatorname{score}_{\Theta}(\mathbf{t}^{(i)}, \mathbf{x}^{(i)}))}{\sum_{\mathbf{t}^{(i)}} \exp (\operatorname{score}_{\Theta}(\mathbf{t}^{(i)}, \mathbf{x}^{(i)}))},
\end{equation}
\label{equation 15}
\begin{equation}
\operatorname{score}_{\Theta}(\mathbf{t}^{(i)}, \mathbf{x}^{(i)})=\sum_{l=1}^L(\text{Emission}_{l, t_l^{(i)}}+\text {CrfTransition}_{t_{l-1}^{(i)}, t_l^{(i)}}),
\end{equation}
where $\text{Emission} \in \mathbb{R}^{L \times K}$ is the \textit{emission score matrix} coming from the logit outputs of BERT ($\text{Emission}=f_{\Theta_{\textbf{BERT}}}(\mathbf{x}^{(i)})$), and
CrfTransition $\in \mathbb{R}^{K \times K}$ is the \textit{CRF transition matrix}.
Model parameters are $\Theta=\{\Theta_{\text{BERT}},   \text{CrfTransition}  \}$.
\textbf{(2)}
Similarly, our proposed Neural-Hidden-CRF is also a exponential discriminative model, concerning the weak label sequence $\mathbf{y}^{(i)}$ and label sequence $\mathbf{t}^{(i)}$  given the sentence $\mathbf{x}^{(i)}$:
\begin{equation}
p(\mathbf{y}^{(i)}, \mathbf{t}^{(i)} \mid \mathbf{x}^{(i)}; \Theta)=\frac{\exp (\operatorname{score}_{\Theta}(\mathbf{y}^{(i)}, \mathbf{t}^{(i)}, \mathbf{x}^{(i)}))}{\sum_{\mathbf{y}^{(i)}} \sum_{\mathbf{t}^{(i)}} \exp (\operatorname{score}_{\Theta}(\mathbf{y}^{(i)}, \mathbf{t}^{(i)}, \mathbf{x}^{(i)}))},
\label{equation 16}
\end{equation}
\begin{small}
\begin{equation}
\begin{aligned}
&
\operatorname{score}_{\Theta}(\mathbf{y}^{(i)}, \mathbf{t}^{(i)}, \mathbf{x}^{(i)})=\sum_{l=1}^L(\text {Emission}_{l, t_l^{(i)}}+\text {CrfTransition}_{t_{l-1}^{(i)}, t_l^{(i)}} + \\
&\text{WeskSourceTransition\#1}_{t_{l}^{(i)}, y_l^{(i,1)}} + ... + \text{WeskSourceTransition\#J}_{t_{l}^{(i)}, y_l^{(i,J)}}),
\end{aligned}
\label{equation 17}
\end{equation}
\end{small}
where $\text{Emission}$, $\text{CrfTransition}$ have the same meaning as those in model BERT-CRF above, and each $\text{WeakSourceTransition} \in \mathbb{R}^{K \times K}$ refers to the \textit{weak source transition matrix} introduced in Section~\ref{Feature Functions}.
Model parameters are $\Theta=\{\Theta_{\text{BERT}}, \text{CrfTransition},$ $\text{WeakSourceTransition\#1}, ...,  \text{WeakSourceTransition\#J} \}$.

\subsection{Learning}
\label{Learning}
Given the weak supervision data $\mathcal{D}=\{\mathbf{x}^{(i)}, \mathbf{y}^{(i)}\}_{i=1}^{I}$ and the model constructed above, we estimate the parameters of the model by maximizing the conditional log-likelihood involving the latent ground truth variable:
\begin{equation}
\begin{aligned}
& \mathcal{L}(\Theta)  =\sum_{i} \log p(\mathbf{y}^{(i)} \mid \mathbf{x}^{(i)}; \Theta).
\end{aligned}
\label{equation 18}
\end{equation}
Further, 
\begin{small}
\begin{equation}
\begin{aligned}
& \log p(\mathbf{y}^{(i)} \mid \mathbf{x}^{(i)}; \Theta) \\
=& \log \sum_{\mathbf{t}^{(i)}}   p(\mathbf{y}^{(i)}, \mathbf{t}^{(i)} \mid \mathbf{x}^{(i)}; \Theta) \\
=&\log  \frac{1}{\boldsymbol{Z}(\mathbf{x}^{(i)}; \Theta)} \sum_{\mathbf{t}^{(i)}}  \exp \left(\sum_{l}\sum_{w} \theta_{w} \cdot f_{w}(\mathbf{y}^{(i)}_{l}, t^{(i)}_{l-1}, t^{(i)}_{l},
\mathbf{x}^{(i)}, l)\right),
\end{aligned}
\label{equation 19}
\end{equation}
\end{small}
where instance-specific normalization factor $\boldsymbol{Z}(\mathbf{x}^{(i)}; \Theta)$ defined before is 
\begin{equation}
\boldsymbol{Z}(\mathbf{x}^{(i)}; \Theta)=\sum_{\mathbf{y}^{(i)}} \sum_{\mathbf{t}^{(i)}} \exp \left(\sum_{l}\sum_{w} \theta_{w} \cdot f_{w}(\mathbf{y}^{(i)}_{l},
t^{(i)}_{l-1},
t^{(i)}_{l},
\mathbf{x}^{(i)}, l)\right).
\label{equation 20}
\end{equation}
Calculation of $\log  \sum_{\mathbf{t}^{(i)}}  \exp \left(\sum_{l}\sum_{w} \theta_{w} \cdot f_{w}(\mathbf{y}^{(i)}_{l}, t^{(i)}_{l-1}, t^{(i)}_{l},\mathbf{x}^{(i)}, l)\right)$ and $\log  \boldsymbol{Z}(
\mathbf{x}^{(i)}; \Theta)$---similar to the corresponding calculation of the CRF---can be efficiently solved by dynamic programming algorithm.
The detail derivations are shown in Appendix~\ref{Calculation of the Likelihood}.

The above $\mathcal{L}(\Theta)$ provides a unified objective function for optimization in Neural-Hidden-CRF, which can be done with standard stochastic optimization techniques, such as SGD~\cite{goodfellow2016deep} or Adam~\cite{kingma2014adam}.

\subsection{Inference}
\label{Inference}
At the test phase, given a new test sequence $\mathbf{x}$, we want to infer the most probable ground truth sequence $\mathbf{t}^*=\arg \max _{\mathbf{t}^*} p(\mathbf{t}^* \mid \mathbf{x} ;\Theta)$.
Here we can ignore the parameters of the weak source transition matrix part and use the classifier (e.g., BERT-CRF or BiLSTM-CRF) within Neural-Hidden-CRF to make the inference.
Like the CRFs, this inference problem can be solved efficiently with the canonical Viterbi algorithm \cite{forney1973viterbi}, which applies the dynamic programming.

\subsection{Implementation Details}
\label{Implementation Details}
\paragraph{Parameter initialization}
In our model, similar to the initialization in weak supervision model MAX-MIG~\cite{cao2019max}, we can initialize the parameters of the weak  sources (i.e., the \textit{weak source transition matrix}, denoted as $\mathbf{\Pi}^{(j)}$) as:
\begin{equation}
\pi_{m n}^{(j)}=\rho\cdot\frac{\sum_{i=1}^{I} \sum_{l=1}^{L}\mathbb{I}(t^{(i)}_{l}=m) \mathbb{I}(y^{(i, j)}_{l}=n)}
{\sum_{i=1}^{I} \sum_{l=1}^{L}\mathbb{I}(t^{(i)}_{l}=m) \mathbb{I}(y_{l}^{(i, j)} \neq 0)}\mathrm{,}
\label{equation 21}
\end{equation}
where $\rho$ is a hyper-parameter and $t^{(i)}_{l}$ can be easily obtained by majority voting method.
In addition, for the parameters of the classifier part (i.e., parameters in Neural-Hidden-CRF other than the weak source  transition matrices), we can easily pre-train the classifier using the labels inferred by majority voting to obtain a better  parameter initialization for the model.

\subsection{Others: Computational Complexity}
\label{Others: Computational Complexity}
The computational complexities of our method and some representative methods are shown in Table~\ref{Table:1}, which contains the complexities of (1) performing the probability calculation on likelihood/objective during learning and (2) performing inference. In summary, our method has the same complexities as many existing methods.

\begin{table}[h]
\caption{Computational Complexity. 
$L$/$K$/$J$: sequence length/\# categories/\# weak sources.}
\centering
\scalebox{0.73}{
\begin{threeparttable}  
{
\begin{tabular}{  l c c}
\toprule
Method
&  Probability calculation  & Inference
\tabularnewline
\midrule
\rule{0pt}{7pt}MV + BERT-CRF$^{\S}$
&  $O(LK^2)$  &  $O(LK^2)$ 
\\
\cline{2-3}
\rule{0pt}{9pt}LSTM-Crowd~\cite{nguyen2017aggregating}$^{*}$, LSTM-Crowd-cat~\cite{nguyen2017aggregating}$^{*}$
& 
\multirow{2}{*}{$O(JLK^2)$} & 
\multirow{2}{*}{$O(LK^2)$ } \\
CONNET~\cite{lan2019learning}$^{*}$, Zhang et al.~\cite{zhang2021crowdsourcing}$^{*}$
& &     \\
\cline{2-3}
\rule{0pt}{10pt}Ours$^{*}$
  &   $O(JLK^2)$  &  $O(LK^2)$ 
\tabularnewline
\bottomrule
\end{tabular}
}
\begin{tablenotes}
 \footnotesize   
\item[1] $\S$: 
When we use the labels inferred from a truth inference method and perform supervised training, e.g., MV+BERT-CRF, the complexities of  MV+BERT-CRF are the same as CRF~\cite{collins2015log}.
\item[2] $*$: 
Our method has the same complexities as the ``source-specific perturbation'' methods LSTM-Crowd~\cite{nguyen2017aggregating}, LSTM-Crowd-cat~\cite{nguyen2017aggregating}, CONNET~\cite{lan2019learning} and Zhang et al.~\cite{zhang2021crowdsourcing}. 
This is because: 
(\textit{i}) For the probability calculation complexity, since the ``source-specific perturbation'' methods require to learn $J$ source-specific models, their complexities ($O(JLK^2)$) are $J$ times the corresponding complexity of CRF ($O(LK^2)$). Also, when our method utilizes the dynamic programming algorithm to compute our Equation~\ref{A:3} (the most significant consumers of computing) in the Appendix~\ref{Calculation of the Likelihood}, its complexity is also $O(JLK^2)$;
(\textit{ii}) For the inference complexity, each Viterbi decoding process required by these methods is the same as for the CRF, and therefore the complexities are all $O(LK^2)$.
\item[3]
Note that here we consider the complexities on one instance, and by the general convention, we do not consider the complexity arising from the deep neural backbone, which has an equivalent effect for all methods.
\end{tablenotes}
\end{threeparttable} 
}
\smallskip
\label{Table:1}
\hfil
\captionsetup{type=table,skip=5pt}
\end{table}

\section{Experiments}
\label{experiments}

\subsection{Setup}

\subsubsection{Datasets}
\label{Datasets}
We evaluate the proposed Neural-Hidden-CRF on four widely-used, publicly available WS datasets, including the CoNLL-03 (MTurk) dataset~\cite{rodrigues2014sequence,rodrigues2018deep} contributed by crowdsourcing workers from Amazon Mechanical Turk (MTurk)\footnote{\url{https://www.mturk.com/}}, and three datasets~\cite{zhang2021wrench} (CoNLL-03 (WS),  WikiGold (WS), MIT-Restaurant (WS)) labeled from  artificially pre-defined label functions.
Table~\ref{Table:2} shows the main statistics.
Specifically:
\textbf{(1) CoNLL-03 (MTurk)}~\cite{rodrigues2014sequence,rodrigues2018deep} is constructed on the well-established CoNLL-03 dataset~\cite{sang2003introduction} through introducing additional crowdsourcing annotations.
The goal is to recognize named entities (\textit{person, location, organization, miscellaneous}) together with their different parts (\textit{begin, inside}) in the sentence. 
We shuffled and divided the original $3250$ test samples in ~\citet{rodrigues2014sequence} into a validation set and a test set containing 2000/1250 samples, respectively;
\textbf{(2) CoNLL-03 (WS), WikiGold (WS) and MIT-Restaurant (WS)}
are utilized and open-sourced in the recently proposed WS benchmark called Wrench~\cite{zhang2021wrench,rodrigues2018deep}.
These three datasets cover three different domains, and detailed information about them is provided in~\citet{zhang2021wrench}.
\begin{table}[h]
\caption{Statistics of all datasets.}
\centering
\scalebox{0.758}{
\fontsize{1pt}{7pt}
\begin{threeparttable}  
\setlength{\tabcolsep}{0.5mm}{
\begin{tabular}{  l l r r r}
\toprule
 Dataset
&  Domain
  & \#Data(train/val/test)
   & \#Entities
   & \#Source  
 \tabularnewline
\midrule
CoNLL (MTurk)
& News &  5,985/2,000/1,250  & 4 & 47
\\
CoNLL (WS)& News  & 14,041/3,250/3,453   &4 &16   \\
WikiGold (WS)
& Web Text &  1,355/169/170  & 4 & 16\\
MIT-Rest. (WS)
  &  Review & 7,159/500/1,521 & 8  &16
\tabularnewline
\bottomrule
\end{tabular}
}
\end{threeparttable} 
\smallskip
}
\label{Table:2}
\hfil
\captionsetup{type=table,skip=5pt}
\end{table}

\subsubsection{Compared Methods.}
\textbf{(1) On CoNLL-03 (MTurk).}
We consider the following  methods:
(\textit{i}) 
MV-BiLSTM/MV-BiLSTM-CRF: They are the two-stage learning baselines, which first estimate the ground truth from weak labels by MV (Majority Voting), and then train the LSTM/LSTM-CRF;
(\textit{ii}) 
CL (VW), CL (VW+B) and CL (MW): They are three variants of the representative WSSL method Crowd-Layer~\cite{rodrigues2018deep}, where ``VW'', ``VW+B'' and ``MW'' refer to  three different ways of parameterizing weak source reliability; 
(\textit{iii}) 
LSTM-Crowd~\cite{nguyen2017aggregating}, LSTM-Crowd-cat~\cite{nguyen2017aggregating}, \citet{zhang2021crowdsourcing}, and CONNET~\cite{lan2019learning}: These four methods, which apply the ``source-specific perturbation'' mentioned in Section~\ref{introduction}, dominate the deep learning-based WSSL methods and show the competitive results~\cite{lan2019learning};
(\textit{iv}) 
OptSLA~\cite{sabetpour2020optsla} and AggSLC~\cite{sabetpour2021truth}: They both follow the approach of constructing an optimization objective containing weak source weights, classifier parameters, latent ground truth, and iteratively updating them using a coordinate ascent algorithm;  
(\textit{v}) 
CRF-MA~\cite{rodrigues2013learning}: This is a partial directed graphical model where the ground truth sequence is also modeled as a latent variable and each weak source's behavior pattern is modeled by a specific scalar;
(\textit{vi}) 
HMM-Crowd~\cite{nguyen2017aggregating} and BSC-seq~\cite{simpson2018bayesian}: They belong to the HMM-based graphical models mentioned in Section~\ref{introduction}, where the latter is a Bayesian version of the former;
(\textit{vii}) 
Finally, we consider Gold, denoting the classifier (BiLSTM-CRF) trained in the ideal case when true labels are known.
\textbf{(2) On CoNLL-03 (WS), WikiGold (WS) and MIT-Restaurant (WS).}
We compared many methods by using the results reported from benchmark Wrench~\cite{zhang2021wrench}.
Specifically, they involves the advanced CONNET~\cite{lan2019learning}, CHMM~\cite{li2021bertifying}, the HMM-based graphical model called HMM~\cite{lison2020named}, and the label models (WMV~\cite{zhang2021wrench}, DS~\cite{dawid1979maximum}, DP~\cite{ratner2016data}, MeTal~\cite{ratner2019training}, FS~\cite{fu2020fast}) for classification task with certain adaptations.

\subsubsection{Configurations.}
The hyper-parameter settings are shown in Appendix~\ref{Experimental Configurations}.
(Also, note that some suggestions for setting hyper-parameters are provided in Appendix~\ref{Suggestions for Setting Hyperparameters}.)
Further:
\textbf{(1) On CoNLL-03 (MTurk).}
We applied the canonical BiLSTM-CRF~\cite{ma2016end}\footnote{We used the publicly available implementation:
\url{https://github.com/ZubinGou/NER-BiLSTM-CRF-PyTorch}.} as the classifier backbone of our model and comparison methods.
\textbf{(2) On CoNLL-03 (WS), WikiGold (WS) and MIT-Restaurant (WS).}
Our experiments on these datasets build on the recent great benchmark Wrench~\cite{zhang2021wrench}, where we adhered rigorously to their various settings and used their open-source code as the foundation for implementing our method.
We used the more advanced language model BERT of the two available choices (BiLSTM and BERT) provided by Wrench as the backbone.

\subsection{Results and Analysis}
\subsubsection{Main Results}
Tables~\ref{Table:5} and \ref{Table:6}---concerning the CoNLL-03 (MTurk) dataset and the other three WS datasets, respectively---show the prediction performance of all methods on the test data and the inference performance on the training/test data, i.e., the performance of inferring the latent ground truth.\footnote{It is worth noting that, unlike the metrics of ``inference on train data'' in Table~\ref{Table:5} and consistent with the approach in the WS benchmark~\cite{zhang2021wrench}, we report in Table~\ref{Table:6} the inference performance of all methods on the test data, where weak labels are also available.}
First, \textit{we find that our model Neural-Hidden-CRF substantially outperforms all the comparison methods by a large margin on the most important average F1 metric on dataset CoNLL-03 (MTurk) and the other three datasets.}
The more robust performance demonstrated by our Neural-Hidden-CRF relative to the SOTA neuralized HMM-based CHMM~\cite{li2021bertifying} largely showcases the effectiveness of our model in leveraging the global optimization perspective offered by the undirected graphical model.
Further and more specifically, on the average F1 metric, Neural-Hidden-CRF outperforms the recently proposed AggSLC~\cite{sabetpour2021truth} by $4.81$  points on  CoNLL-03 (MTurk), and exceeds the SOTA method CHMM~\cite{li2021bertifying} by $2.80/2.23$ points on the three WS datasets.
It is also worth noting that the comparison methods~\cite{nguyen2017aggregating, zhang2021crowdsourcing, lan2019learning, sabetpour2021truth} on CoNLL-03 (MTurk) dataset, apply either the same backbone (i.e., the GloVe 100-dimensional word embeddings along with BiLSMT-CRF in~\citet{nguyen2017aggregating}) as ours, or more advanced backbones (i.e., BERT-BiLSTM-CRF  in~\citet{zhang2021crowdsourcing}, Efficient ELMO along with BiLSTM-CRF in~\citet{lan2019learning}, BERT in AggSLC~\cite{sabetpour2021truth}) than ours.

\begin{table*}[t]
\centering
\caption{Performance  ($\%$)  on  
 CoNLL-03  (MTurk) dataset. Results  are averaged over 20 runs.
The best results under the F1 metric of most interest are marked in \textbf{bold}.}
\scalebox{0.85}{
\begin{threeparttable}  
\begin{tabular}
{l lcccc cc
>{\columncolor{lightgray!20}}c}
\toprule  \multicolumn{2}{c}{} & \multicolumn{3}{c}{\textbf{Prediction on test data}$^{\S}$} & \multicolumn{3}{c}{\textbf{Inference on train data}$^{*}$ } & \multicolumn{1}{c}{}  \\
  \cmidrule(lr){3-5} 
  \cmidrule(lr){6-8}
\textbf{Paradigm} &
\textbf{Method}  & \textbf{Precision}       & \textbf{Recall}        & \textbf{F1}           & \textbf{Precision}      
              & \textbf{Recall}   &              \textbf{ F1}
                          & \textbf{\underline{Avg. F1}}
\tabularnewline
\midrule
\multirow{2}{*}{Two-stage WSSL} & 
 MV + BiLSTM-CRF & 87.19($\pm$1.19)  &  65.00($\pm$3.28)    & 74.41($\pm$2.11)  &  86.27($\pm$1.08)  &  66.06($\pm$2.3) 
 & 74.79($\pm$1.38) &  74.60
\tabularnewline
& 
 MV + BiLSTM & 82.21($\pm$1.46)   &  61.30($\pm$2.57)        & 70.20($\pm$1.69)   &  80.62($\pm$1.01)  &  61.82($\pm$2.36) 
 & 69.96($\pm$1.64)   &  70.08
\tabularnewline
\midrule
\multirow{11}{*}{One-stage WSSL} 
& 
 CL (VW)~\cite{rodrigues2018deep}  & 83.93($\pm$0.83)  &  61.50($\pm$2.07)        & 70.96($\pm$1.46)  &  82.90($\pm$0.71)  &  64.02($\pm$1.76)
 & 72.24($\pm$1.29)   &  71.60
\tabularnewline
& 
 CL (VW+B)~\cite{rodrigues2018deep}    & 81.93($\pm$1.57)   &  61.00($\pm$2.89)       & 69.87($\pm$1.62)  &  80.31($\pm$1.38) &  61.70($\pm$2.65)
 & 69.75($\pm$1.73) &  69.81
\tabularnewline
& 
 CL (MW)~\cite{rodrigues2018deep}
   & 83.93($\pm$0.89)  &  61.33($\pm$1.65)       & 70.86($\pm$1.65)
  &  82.24($\pm$0.55) &  62.91($\pm$1.26)
 & 71.27($\pm$0.88)
 &   71.07
  \tabularnewline
  &
 LSTM-Crowd~\cite{nguyen2017aggregating}$^{\dag}$   & 82.38  &  62.10  &  70.82 &   -   &  -   &   -   & -
\tabularnewline
 &
LSTM-Crowd-cat~\cite{nguyen2017aggregating}$^{\dag}$   &    79.61    & 62.87  &  70.26 &  -  &  -  & -  &  -
\tabularnewline
 &
\citet{zhang2021crowdsourcing}$^{\dag}$ 
&  78.84  &  75.67  &  77.95
&  -   &  -   &   - &  -
\tabularnewline
 &
CONNET~\cite{lan2019learning}$^{\dag}$  & 87.77($\pm$0.25) & 72.79($\pm$0.04) & 79.99($\pm$0.08) &  - &  -  &    -   &  -
\tabularnewline

 &
AggSLC~\cite{sabetpour2021truth}$^{\dag}$     & 70.95 & 77.16 &  73.93 &  83.02   &  78.69   &   80.79 &  77.36
\tabularnewline
 &
CRF-MA~\cite{rodrigues2014sequence}$^{\dag}$     & 49.4 & 85.6 & 62.6  &  86.0   &  65.6   &   74.4  &  68.5
\tabularnewline
 \cmidrule(lr){2-9} 
 & 
\textbf{Neural-Hidden-CRF}
     &  82.25($\pm$1.05)   &  80.93($\pm$1.05)    &  \textbf{82.06}($\pm$0.63)
 &  84.41($\pm$1.04) &  80.28($\pm$0.74) & 
\textbf{82.28}($\pm$0.49) &  \textbf{82.17}
\tabularnewline
\midrule
\multirow{4}{*}{Truth Inference} & 
MV   &  -   &  -     &  -  & 79.12($\pm$0.00) & 58.50($\pm$0.00)   & 67.27($\pm$0.00)  &  -
\tabularnewline
 &
OptSLA~\cite{sabetpour2020optsla}$^{\dag}$  &  -   &  -     &  -  &  79.42 &   77.59  & 78.49 &  -
\tabularnewline
 &
HMM-Crowd~\cite{nguyen2017aggregating}$^{\dag}$   &  -   &  -     &  -  & 77.40 &  72.29   & 74.76 &  -
\tabularnewline
 &
BSC-seq~\cite{simpson2018bayesian}$^{\dag}$  &  -   &  -     &  -  & 80.3 &  74.8   & 77.4 &  -
\tabularnewline
\midrule
-
& 
\gc{Gold (Upper Bound)}     &  \gc{91.94($\pm$0.66)}   &   \gc{91.49($\pm$0.87)}      &  \gc{91.71($\pm$0.75)}  &  \gc{100} &  \gc{100}   & \gc{100} &  \gc{95.86} 
\tabularnewline
\bottomrule
\end{tabular}
\begin{tablenotes}
\footnotesize   
\item[1] $\S$/$*$: Learn from weak supervision labels on the train data and predict on the test data/learn from weak supervision labels on the train data and infer the latent ground truth labels.
\item[2] $\dag$:
Results are reported from the original works.
Note that there are some blanks in these results, as most of these methods reported one of two metrics in their original works.
\end{tablenotes}
\end{threeparttable} 
}
\smallskip
\label{Table:5}
\hfil
\captionsetup{type=table,skip=5pt}
\end{table*}

\begin{table*}[t]
\centering
\caption{Performance  ($\%$)  on WS benchmark datasets from~\citet{zhang2021wrench}.
Our results  are averaged over 20 runs.
The best results  are marked in \textbf{bold}.
Each table cell contains F1 score with standard deviation and (Precision, Recall) in the bracket.}
\scalebox{0.85}{
\begin{threeparttable}  
\begin{tabular}
{l lcccc cc
>{\columncolor{lightgray!20}}c}
 \toprule 
         \multicolumn{2}{c}{} & \multicolumn{3}{c}{\textbf{Prediction on test data$^{\S}$}} &
    \multicolumn{3}{c}{\textbf{Inference on test data$^{*}$}} & \multicolumn{1}{c}{}  \\
  \cmidrule(lr){3-5} 
  \cmidrule(lr){6-8}
\textbf{Paradigm} &
\textbf{Method}
      & \textbf{CoNLL-03}
              & \textbf{WikiGold}
                        & \textbf{MIT-Rest.}
                            & \textbf{CoNLL-03}      
              & \textbf{WikiGold}
                         & \textbf{MIT-Rest.}
                          & \textbf{\underline{Avg.F1(P/I)}}
\tabularnewline
\midrule
\multirow{16}{*}{Two-stage WSSL} & 
\multirow{2}{*}{MV + BERT-CRF~\cite{zhang2021wrench}$^{\dag}$} & 66.63($\pm$0.85)  &  62.09($\pm$1.06)       & 42.95($\pm$0.43)  
& 60.36($\pm$0.00) & 52.24($\pm$0.00)  & \textbf{48.71}($\pm$0.00)
  &  57.22/53.77
\tabularnewline
&  & \gc{(67.68/65.62)}  &  \gc{(61.89/62.29) }      & \gc{(63.18/32.54)} 
& \gc{(59.06/61.72)}  & \gc{(48.95/56.00)} 
 & \gc{(74.25/36.24)}   &  -
\tabularnewline
  \cmidrule(lr){2-9} 
&  \multirow{2}{*}{WMV + BERT-CRF~\cite{zhang2021wrench}$^{\dag}$} & 64.38($\pm$1.09) &  59.96($\pm$1.08)       & 42.62($\pm$0.23)
&  60.26($\pm$0.00) &   52.87($\pm$0.00)  & 48.19($\pm$0.00)
 &   55.65/53.77
\tabularnewline
&  & \gc{(66.55/62.35)}  & \gc{(60.33/59.73)}        & \gc{(63.56/32.06)}  
& \gc{(59.03/61.54)}   &  \gc{(50.74/55.20)} 
 & \gc{(73.73/35.80)}   &  -
\tabularnewline
  \cmidrule(lr){2-9} 
&  \multirow{2}{*}{DS + BERT-CRF~\cite{dawid1979maximum}$^{\dag}$} & 53.89($\pm$1.42)  &  48.89($\pm$1.59)       & 42.26($\pm$0.78) 
& 46.76($\pm$0.00) &  42.17($\pm$0.00)  & 46.81($\pm$0.00) 
 &   48.35/42.25
\tabularnewline
&  & \gc{(54.10/53.68)}  &  \gc{(46.80/51.20)}       & \gc{(62.65/31.89)} 
& \gc{(45.29/48.32)}   &  \gc{(40.05/44.53)} 
 & \gc{(71.71/34.75)} &  -
\tabularnewline
  \cmidrule(lr){2-9} 
&  \multirow{2}{*}{DP + BERT-CRF~\cite{ratner2016data}$^{\dag}$} & 65.48($\pm$0.37)  &  61.09($\pm$1.53)      & 42.27($\pm$0.53) 
& 62.43($\pm$0.22) &  54.81($\pm$0.13)   & 47.92($\pm$0.00)
 &   56.28/55.05
\tabularnewline
&  & \gc{(66.76/64.28)}  &  \gc{(61.07/61.12)}       & \gc{(62.81/31.86)} 
 &  \gc{(61.62/63.26)} &  \gc{(53.10/56.64)}
 & \gc{(73.24/35.61)}  &  -
\tabularnewline
  \cmidrule(lr){2-9} 
&  \multirow{2}{*}{MeTal + BERT-CRF~\cite{ratner2019training}$^{\dag}$} & 65.11($\pm$0.69)  &  58.94($\pm$3.22)      & 42.26($\pm$0.49) 
& 60.32($\pm$0.08) &  52.09($\pm$0.23)   & 47.66($\pm$0.00)
  &   55.44/53.37
\tabularnewline
&  & \gc{(66.87/63.45)}  &  \gc{(61.53/56.75)}       & \gc{(62.82/31.84)} 
&  \gc{(59.07/61.63)} &  \gc{(50.31/54.03)}
 & \gc{(73.40/35.29)}
  & -
\tabularnewline
  \cmidrule(lr){2-9} 
&  \multirow{2}{*}{FS + BERT-CRF~\cite{fu2020fast}$^{\dag}$} & 67.34($\pm$0.75)  &  66.44($\pm$1.40)       & 13.80($\pm$0.23)  
& 62.49($\pm$0.00) &  58.29($\pm$0.00)   & 13.86($\pm$0.00)
  &   49.19/44.88
\tabularnewline
&  & \gc{(70.05/64.83)}  &  \gc{(72.86/61.17)}       & \gc{(72.63/7.62)} 
& \gc{(63.25/61.76)}  &  \gc{(62.77/54.40)}
 & \gc{(84.20/7.55)} &  -
\tabularnewline
  \cmidrule(lr){2-9} 
&  \multirow{2}{*}{HMM + BERT-CRF~\cite{lison2020named}$^{\dag}$} & 67.49($\pm$0.89)  &  63.31($\pm$1.02)       & 39.51($\pm$0.72)  
& 62.18($\pm$0.00) &  56.36($\pm$0.00)  & 42.65($\pm$0.00)
  &   56.77/53.73
\tabularnewline
&  & \gc{(71.26/64.14)}  &  \gc{(70.95/57.33)}       & \gc{(62.49/28.90)}  
&  \gc{(66.42/58.45)} &  \gc{(61.51/52.00)}
 & \gc{(71.44/30.40)} 
 &  -
\tabularnewline
  \cmidrule(lr){2-9} 
&  \multirow{2}{*}{CHMM + BERT-CRF~\cite{li2021bertifying}$^{\dag}$} & 66.72($\pm$0.41)  &  63.06($\pm$1.91)       & 42.79($\pm$0.22) 
& 63.22($\pm$0.26) &  58.89($\pm$0.97)   & 47.34($\pm$0.57)
  &   57.52/56.48
\tabularnewline
&  & \gc{(67.17/66.27)}  &  \gc{(62.12/64.11)}       & \gc{(63.19/32.35)}  
&  \gc{(61.93/64.56)} &  \gc{(55.71/62.45)}
 & \gc{(73.05/35.02)}  &  -
\tabularnewline
\midrule

\multirow{4}{*}{One-stage WSSL} 
 &
\multirow{2}{*}{CONNET~\cite{lan2019learning}$^{\dag}$}  &    67.83($\pm$0.62)    &   64.18($\pm$1.71)  &  42.37($\pm$0.72) &  - &  -  &    -   &  58.13/-
\tabularnewline
&  & \gc{(69.37/66.40)}  &  \gc{(72.17/57.92)}       & \gc{(62.88/31.95)}  &  - &  -
 & -  &  -
\tabularnewline
  \cmidrule(lr){2-9} 
& 
\multirow{2}{*}{\textbf{Neural-Hidden-CRF}}
     &  \textbf{69.16}($\pm$0.92)   & \textbf{66.87}($\pm$1.79)
         &  \textbf{44.94}($\pm$0.99)
 &  \textbf{67.99}($\pm$0.58)   & \textbf{59.69}($\pm$0.68)  & 
48.44($\pm$0.86) &   \textbf{60.32}/\textbf{58.71}
\tabularnewline
&  & \gc{(73.13/65.64)}  &  \gc{(73.00/61.87)}       & \gc{(58.27/36.66)}  &  \gc{(73.12/63.55)} &  \gc{(71.23/51.44)}
 & \gc{(68.17/37.85)} &  -
\tabularnewline
\midrule
\multirow{2}{*}{-}
& 
\multirow{2}{*}{\gc{Gold + BERT-CRF$^{\dag}$}}     &  \gc{87.38($\pm$0.34)}   &   \gc{86.78($\pm$0.84)}      &  \gc{78.83($\pm$0.44)}  &  \gc{100.00($\pm$0.00)} &  \gc{100.00($\pm$0.00)}   & \gc{100.00($\pm$0.00)} &  \gc{84.33/100.00} 
\tabularnewline
&  & \gc{(87.70/87.06)} &  \gc{(87.27/86.29)} & 
\gc{(79.14/78.53)}
&  \gc{(100.00/100.00)} & \gc{(100.00/100.00)}
 & \gc{(100.00/100.00)}  &  \gc{-} 
\tabularnewline
\bottomrule
\end{tabular}
\begin{tablenotes}
 \footnotesize   
 \item[1] $\S$/$*$: Learn from weak supervision labels on the train data and predict on the test data/directly learn from weak supervision labels available on the test data and infer the ground truth labels.
\item[2] $\dag$: Results are reported from~\citet{zhang2021wrench}.
\end{tablenotes}
\end{threeparttable} 
}
\smallskip
\label{Table:6}
\hfil
\captionsetup{type=table,skip=5pt}
\end{table*}

\begin{figure*}[t] 
	\centering 
	\vspace{-0.0cm} 
	\subfigtopskip=-0pt 
	\subfigbottomskip=0pt 
	\subfigcapskip=-5pt 
  	\subfigure[Dataset ConNLL-03 (MTurk)]{
		\label{level.sub.3}
		\includegraphics[width=0.5\linewidth]{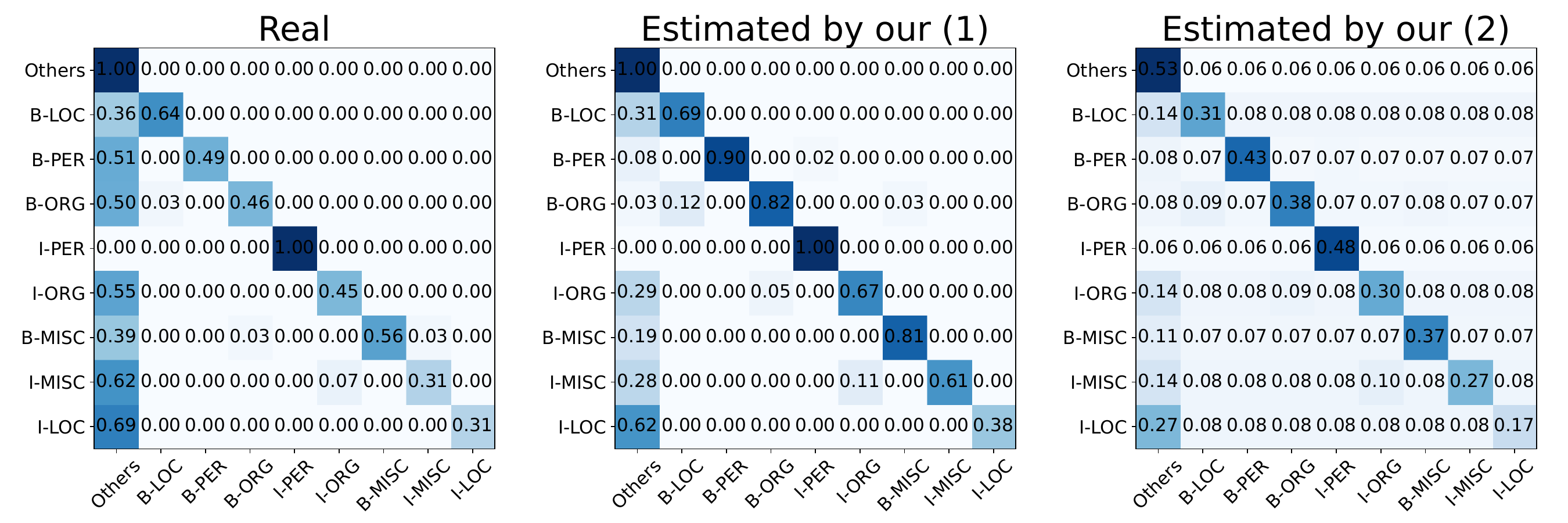}}
        \hspace{6mm}
	\subfigure[Dataset ConNLL-03 (MTurk)]{
		\label{level.sub.4}
		\includegraphics[width=0.2\linewidth]{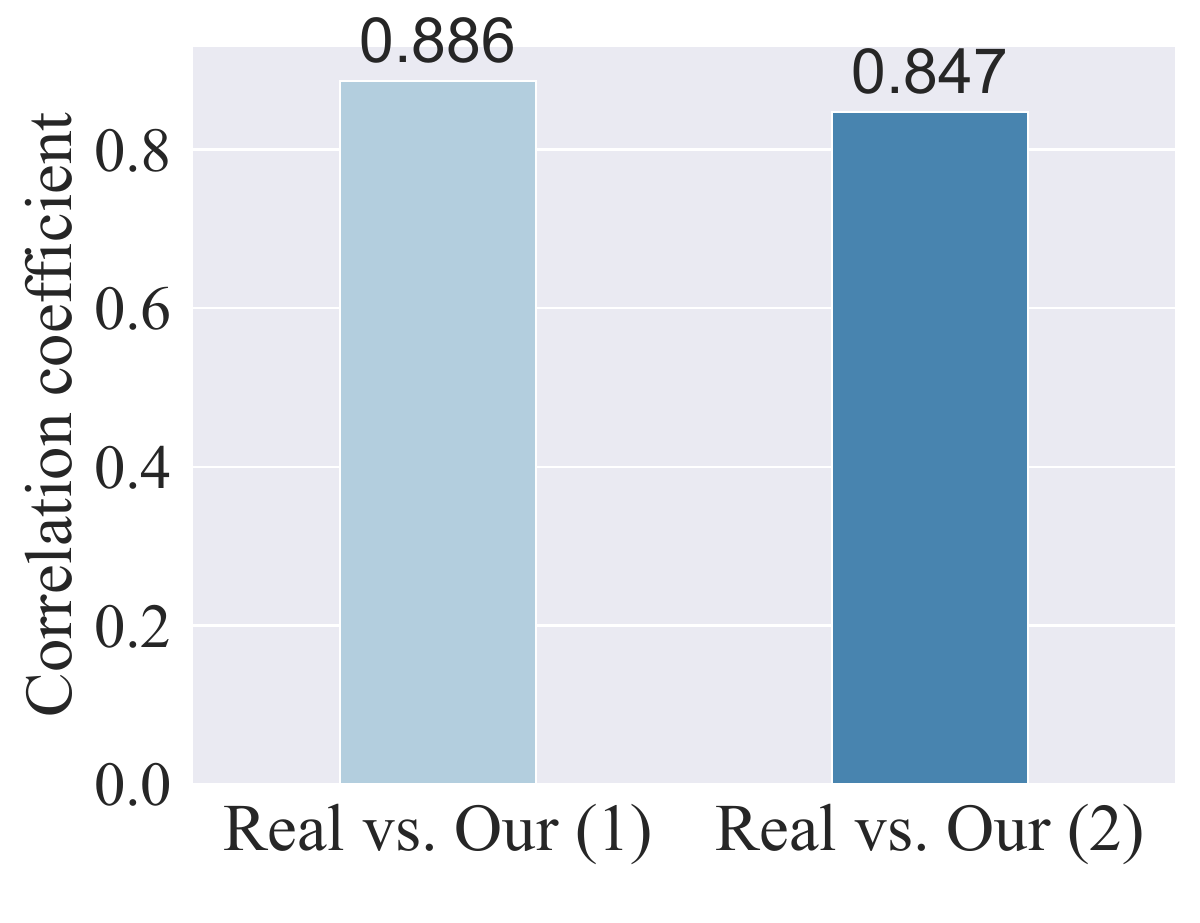}}

	\subfigure[Dataset ConNLL-03 (WS)]{
		\label{level.sub.3}
		\includegraphics[width=0.5\linewidth]{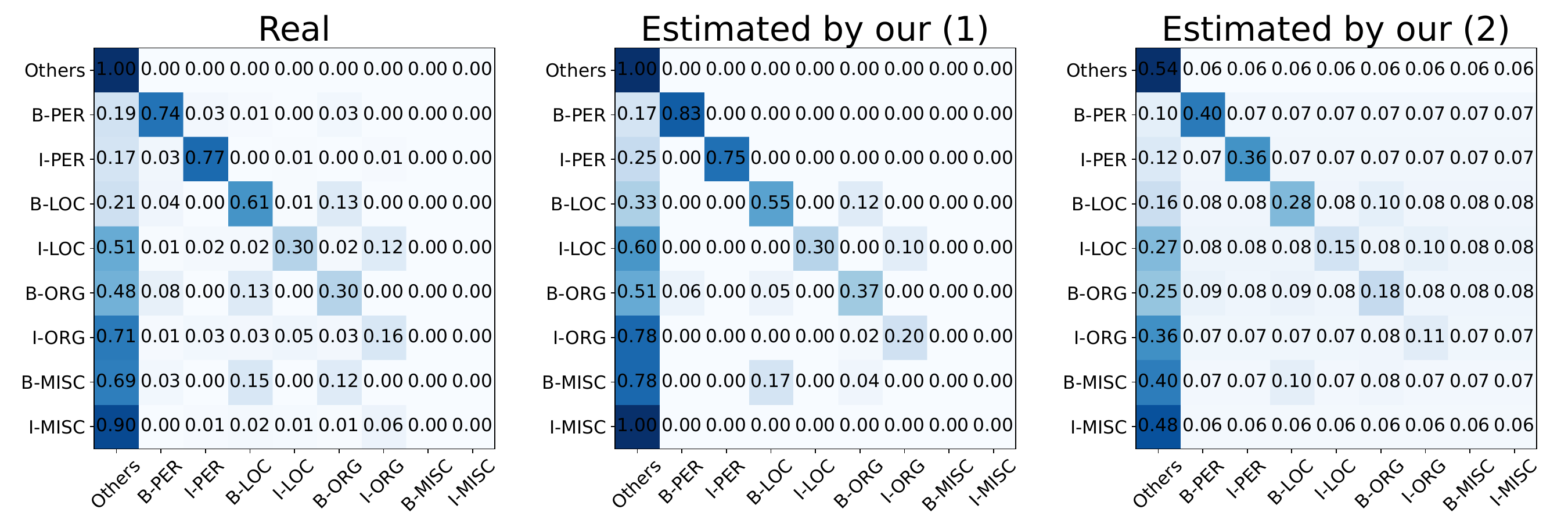}}
  \hspace{6mm}
	\subfigure[Dataset ConNLL-03 (WS)]{
		\label{level.sub.4}
		\includegraphics[width=0.2\linewidth]{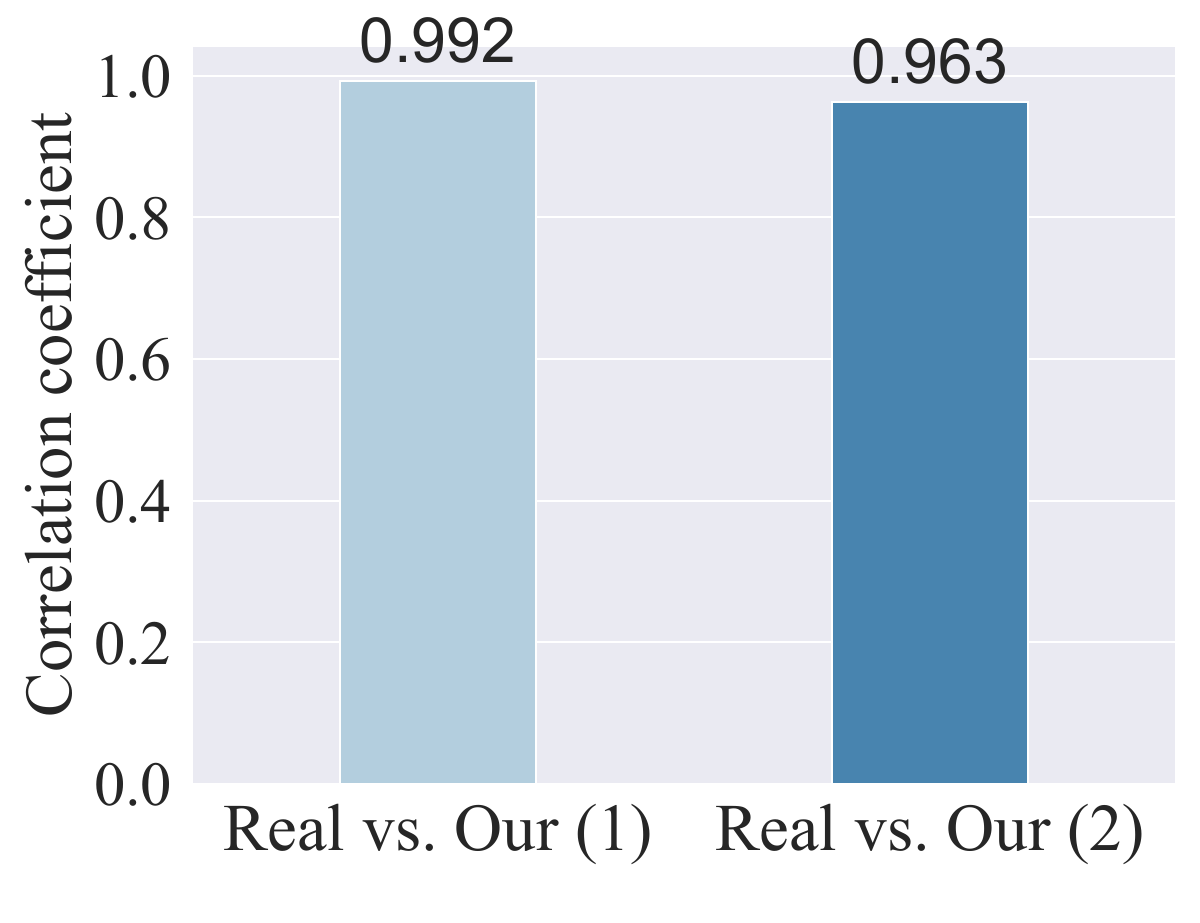}}
	\subfigure[Dataset WikiGold (WS)]{
		\label{level.sub.3}
		\includegraphics[width=0.5\linewidth]{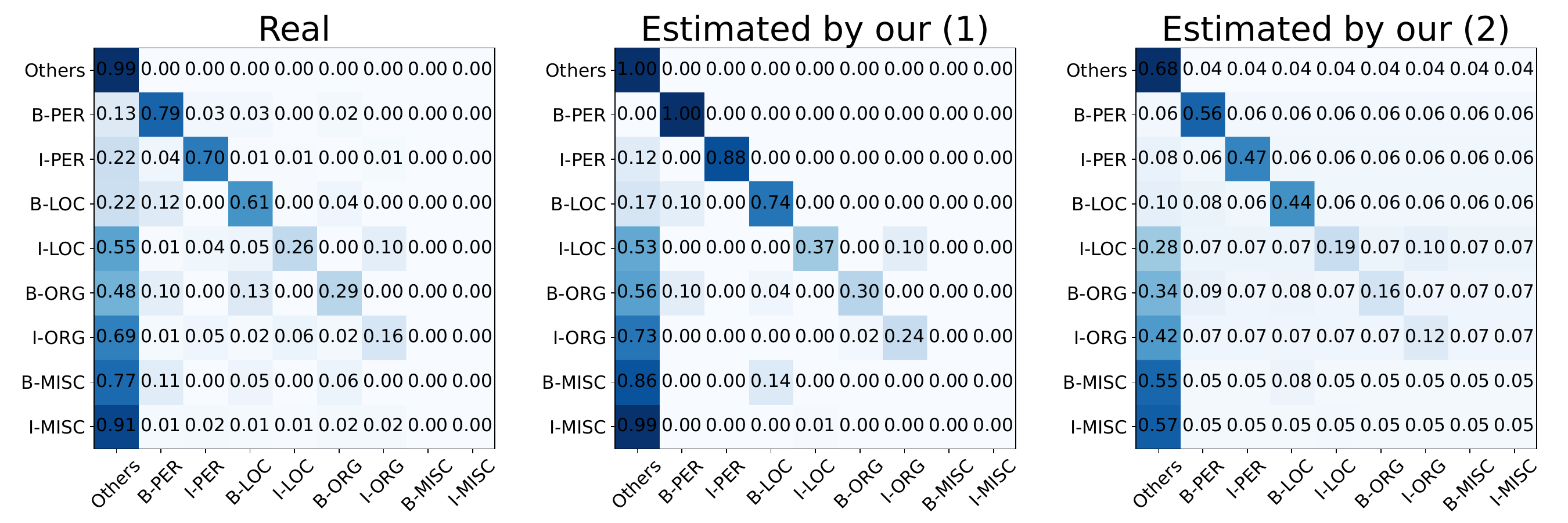}}
  \hspace{6mm}
	\subfigure[Dataset WikiGold (WS)]{
		\label{level.sub.4}
		\includegraphics[width=0.2\linewidth]{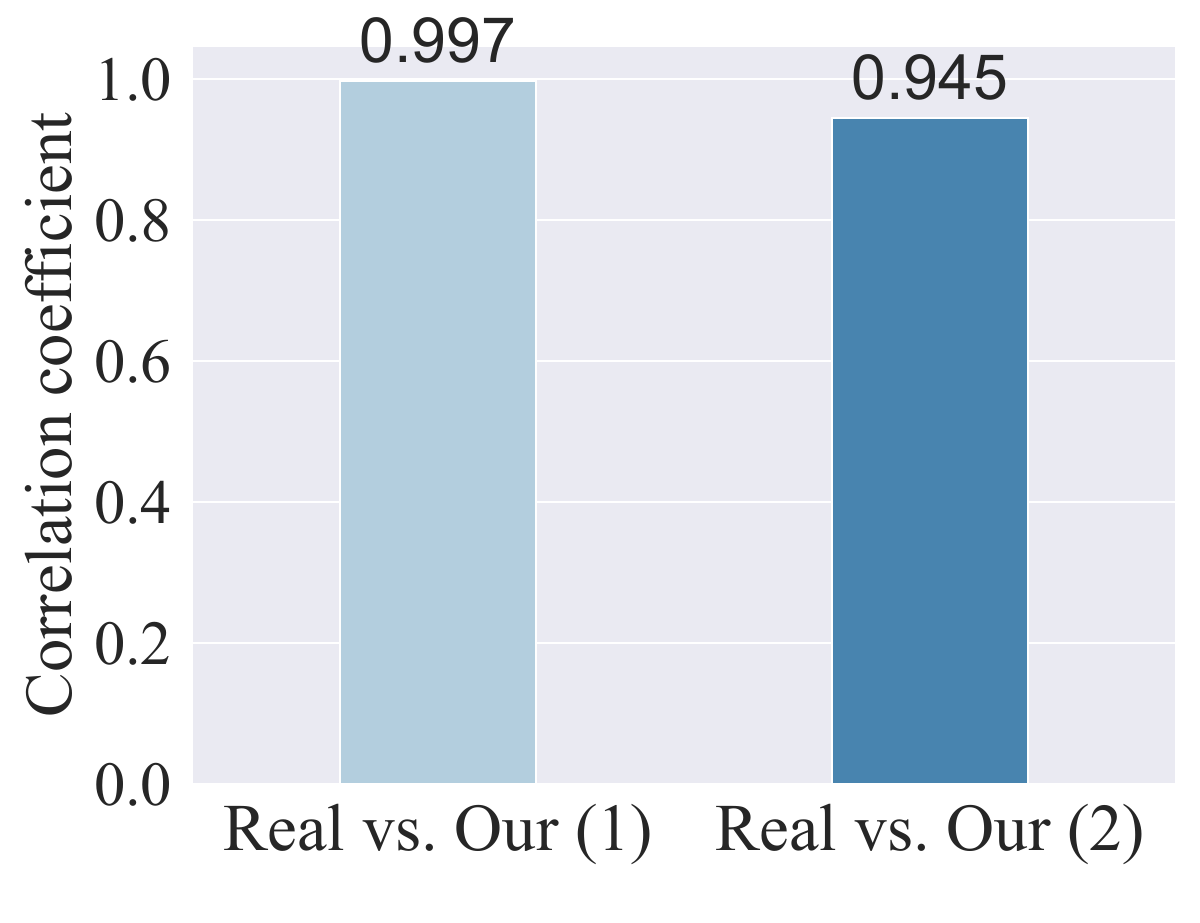}}
  	\subfigure[Dataset MIT-Restaurant (WS)]{
		\label{level.sub.3}
		\includegraphics[width=0.5\linewidth]{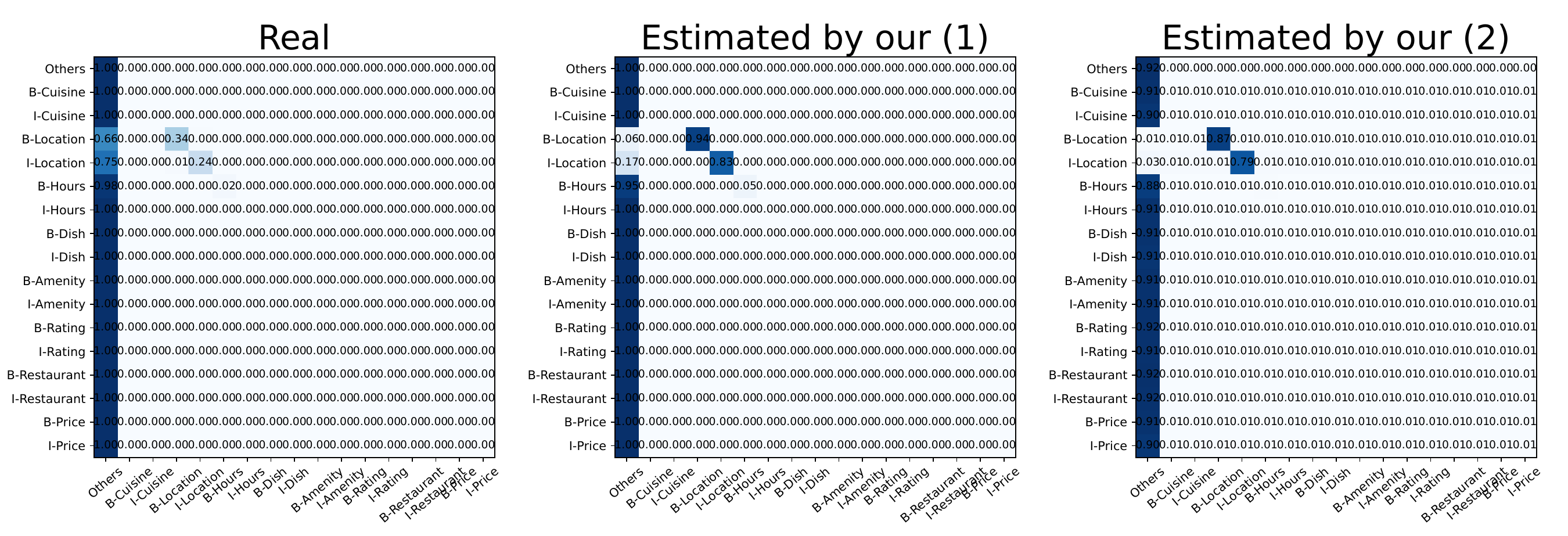}}
  \hspace{6mm}
	\subfigure[Dataset MIT-Restaurant (WS)]{
		\label{level.sub.4}
		\includegraphics[width=0.2\linewidth]{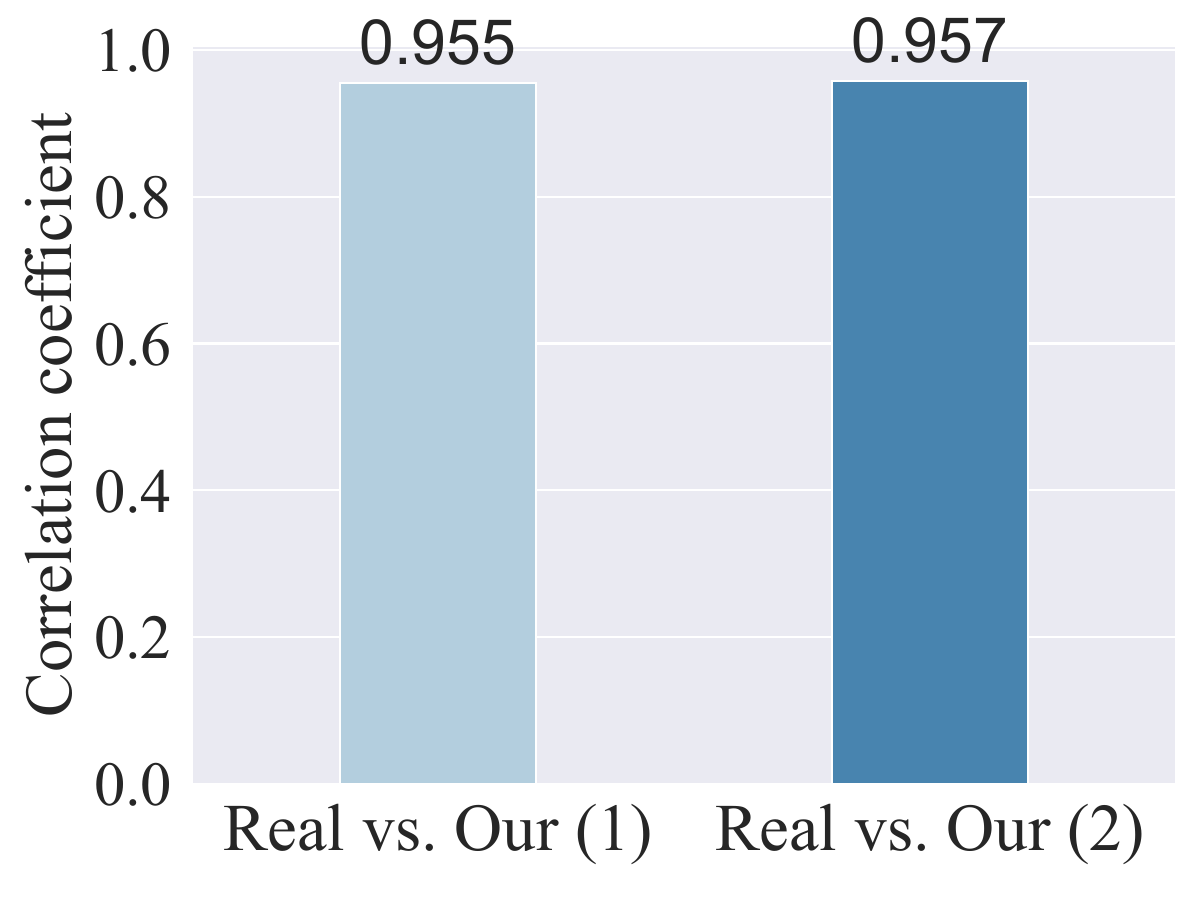}}
 \caption{Comparison between the real weak source  transition matrix and the two weak source transition matrices estimated by Neural-Hidden-CRF on the four datasets.
(\textit{i})
Real: it denotes the probabilistic confusion matrix that we compute by using the truth labels and weak labels in the dataset (in fact, we cannot obtain the real matrix under the theory of our model);
(\textit{ii}) 
Our (1):  it is obtained by setting all non-positive elements of the weak source transition matrix to $0$ and normalizing the elements on each row;
(\textit{iii}) 
Our (2):  it is obtained by exponentiating all elements of the weak source transition matrix and normalizing the elements on each row;
(\textit{iv}) 
In each sub-figure on correlation coefficient, we calculate the element-level values.} 
\label{Figure:3}
\end{figure*}

Compared with the inference metrics, we are more interested in prediction metrics, because in general, our ultimate objective revolves around developing a robust sequence labeler endowed with strong generalization.
For the prediction metric, we find that our Neural-Hidden-CRF outperforms all comparison methods across all datasets, often with considerable margins.
In terms of inference performance, in addition to achieving the second-best result on the MIT-Restaurant (WS) dataset, Neural-Hidden-CRF still often maintains a significant lead on the remaining three datasets and outperforms the second-best by $1.49$/$4.77$/$0.80$ points.

\subsubsection{Weak Source Parameter Estimation and Interpretability}
Benefiting from our use of the interpretable weak source transition matrices in the neuralized undirected graphical model rather than hard-to-interpret neural network parameters to model weak source behavior patterns,  we can now conduct a post-hoc study for the estimated matrices.
Methodologically, for each specific weak supervision source, its parameters can form a matrix of size $K \times K$. 
These parameters possess interpretability on the behavioral pattern of the source. 
That is, the matrix's element at position $(i, j)$ denotes scoring information for the case when the truth is $i$ and the weak label is $j$, where a larger value reflects a greater likelihood.
Empirically, Figure~\ref{Figure:3} shows the results for the respective first weak source on the four datasets.
These results substantiate the accuracy of estimating the weak source transition matrices, and validate that the weak source transition matrices we model do have the \textit{interpretability} in expressing the label transition patterns of weak sources (similar to the CRF transition matrix in the CRF model~\cite{lafferty2001conditional}).
Also, such a result further demonstrates the effectiveness of  Neural-Hidden-CRF from another side.
In additin, the parameters of the weak source transition matrices estimated in Figure~\ref{Figure:3} are unrestricted (i.e., each parameter takes the value space of $(-\infty,+\infty)$), without satisfying probabilistic statutes (as in the HMMs).
Also, our CRF transition matrix is similar.
For example, the elements in the first row of our estimated CRF transition matrix on CoNLL-03 (MTurk) are $[0.24, 2.94, 2.57, 1.69, -3.18, -4.29, 1.11, -3.57, -3.09]$.
These illustrate the resulting flexible scoring comes from the mechanism of holistic undirected graphical modeling and holistic parameter configuration.

\begin{table*}[t]
\centering
\caption{Performance (F1, $\%$) on  ablation study.
Results are averaged over 10 runs.}
\scalebox{0.68}{
\begin{threeparttable}  
\begin{tabular}
{l cccc
>{\columncolor{lightgray!20}}c}
\toprule 
\textbf{Method}
 & \textbf{CoNLL-03(MTurk) (P/I)$^{\S}$}
      & \textbf{CoNLL-03(WS) (P/I/I)$^{*}$}
              & \textbf{WikiGold(WS) (P/I/I)$^{*}$}
                        & \textbf{MIT-Restaurant(WS) (P/I/I)$^{*}$}
                          & \textbf{\underline{Avg.(P/I/I)}$^{*}$}
\tabularnewline
\midrule
W/o-weak-transition & 74.41($\pm$2.11)/74.79($\pm$1.38)   &   66.63($\pm$0.85)/68.61($\pm$0.72)/65.43($\pm$0.51)      &  62.09($\pm$1.06)/60.82($\pm$1.76)/52.32($\pm$0.26) 
 &  42.95($\pm$0.43)/45.00($\pm$0.71)/48.01($\pm$0.73)   &    61.52/62.31/55.25
  \tabularnewline
W/o-crf-transition & 80.79($\pm$0.73)/80.96($\pm$0.23)   &   68.73($\pm$0.71)/70.35($\pm$0.40)/66.78($\pm$0.67)      &  63.89($\pm$1.59)/62.26($\pm$2.14)/58.67($\pm$1.15) 
 &  40.94($\pm$0.86)/42.72($\pm$1.01)/40.24($\pm$4.13)   &    63.59/64.08/55.23
  \tabularnewline
Small-crf-transition & 81.95($\pm$0.70)/82.25($\pm$0.39)   &   69.05($\pm$0.63)/71.25($\pm$0.76)/67.79($\pm$1.13)      &  65.71($\pm$1.68)/64.54($\pm$1.12)/59.38($\pm$1.20) 
 &  42.20($\pm$1.77)/44.19($\pm$1.22)/47.79($\pm$0.62)   &    64.73/65.56/58.32
  \tabularnewline
Small-emission & 68.27($\pm$4.93)/71.20($\pm$4.40)   &   65.99($\pm$1.11)/69.52($\pm$1.53)/64.62($\pm$2.05)      &  61.47($\pm$4.16)/60.57($\pm$2.90)/58.45($\pm$2.78) 
 &  43.48($\pm$1.84)/45.95($\pm$0.64)/47.09($\pm$1.71)   &    59.80/61.81/56.72
  \tabularnewline
Other-classifier-init
 & \textbf{82.43}($\pm$0.64)/82.18($\pm$0.45)   &  69.01($\pm$0.67)/71.66($\pm$0.57)/67.07($\pm$0.84)        &  63.70($\pm$2.99)/63.15($\pm$3.30)/53.61($\pm$0.87)  
 & 42.81($\pm$1.13)/43.95($\pm$1.09)/27.61($\pm$5.63)    &  64.49/65.24/49.43
\tabularnewline
Other-worker-init & 55.15($\pm$10.82)/54.51($\pm$11.35)  & 66.53($\pm$0.74)/68.96($\pm$0.48)/65.42($\pm$0.96)        & 62.40($\pm$1.59)/60.68($\pm$1.47)/53.12($\pm$1.00) 
&  41.57($\pm$0.64)/45.04($\pm$1.00)/39.96($\pm$8.15)    &   56.41/57.30/52.83
\tabularnewline
Other-both-init & 43.00($\pm$13.07)/40.51($\pm$11.60) &  66.40($\pm$1.18)/68.85($\pm$0.97)/65.86($\pm$1.04)      & 63.43($\pm$1.26)/61.88($\pm$1.35)/52.95($\pm$0.81) 
 & 40.55($\pm$0.88)/43.81($\pm$0.89)/36.91($\pm$8.85)   &   53.35/53.76/51.91
\tabularnewline
Freeze-source & 79.75($\pm$1.09)/80.63($\pm$0.26)  &  67.58($\pm$0.80)/70.29($\pm$0.74)/67.46($\pm$0.47)       & 65.70($\pm$1.87)/65.34($\pm$2.08)/58.03($\pm$1.81) 
 & 44.54($\pm$0.35)/46.19($\pm$0.36)/47.04($\pm$0.84)   &   64.39/65.61/57.51
\tabularnewline
\midrule
\textbf{Neural-Hidden-CRF}
    &  82.06($\pm$0.63)/\textbf{82.28}($\pm$0.49)   & \textbf{69.16}($\pm$0.92)/\textbf{71.89}($\pm$0.55)/\textbf{67.99}($\pm$0.58)   
         &  \textbf{66.87}($\pm$1.79)/\textbf{65.55}($\pm$1.33)/\textbf{59.69}($\pm$0.68)  
 &  \textbf{44.94}($\pm$0.99)/\textbf{46.61}($\pm$0.91)/\textbf{48.44}($\pm$0.86)     &   \textbf{65.76}/\textbf{66.58}/\textbf{58.71}
\tabularnewline
\bottomrule
\end{tabular}
\begin{tablenotes}
 \footnotesize   
 \normalsize
 \item[1] $\S$: ``I'' denotes we learn from weak supervision labels on the train data and infer the latent ground truth labels.
 \item[2] $*$:  ``I/I'' denote we learn from weak supervision labels on train/test data and infer the latent ground truth labels on the train/test data, respectively.
Note that the latter three datasets are different from dataset ConLL-03 (MTurk), because they also contain weak supervision labels on the test data.
\end{tablenotes}
\end{threeparttable} 
}
\smallskip
\label{Table:7}
\hfil
\captionsetup{type=table,skip=5pt}
\end{table*}

\subsubsection{Equipped with Other Backbones}
The deep model in our model assumes a backbone role as a feature extractor for sentence sequences.
Theoretically, a more powerful deep model would be more conducive to extracting more useful contextual semantic information and delivering more accurate prediction information about the truth sequences, thus having more potential to improve the final performance.
Here we conducted a small-scale study on  partial datasets, where the obtained  results align with the above analysis.
That is, for the prediction task on datasets CoNLL-03 (WS) and WikiGold (WS), our Neural-Hidden-CRF yields suboptimal F1 performance relative to the original BERT-based one when we apply the relatively weaker deep model BiLSTM (provided by the benchmark Wrench~\cite{zhang2021wrench})---BiLSTM-based/BERT-based: $67.63(\pm1.08)$/$69.16(\pm0.92)$, $65.21(\pm1.45)$/$66.87(\pm1.79)$.
(Settings of \texttt{Batch}/\texttt{Lr}/\texttt{Lr\_weak}/\texttt{\(\rho\)}: $64$/$0.005$/$0.0001$/$3.0$, $32$/$0.001$/$0.0001$/$3.0$.)

\subsubsection{Ablation Study}
Here we consider an extensive array of possible variants, involving 
\textit{the ablation of different components} (variants i-iv),
\textit{the use of different parameter initialization} (variants v-vii), and 
\textit{the freezing of model parameters} (variant viii).
Specifically:
\textbf{(\textit{i})} W/o-weak-transition:
We ablate the weak source transition matrix, where we use the results inferred by the MV (Majority Voting) to represent the latent truth sequence and perform supervised learning, so that the dependencies between the truth sequence $\mathbf{t}$ and the weak label sequence $\mathbf{y}$ are not taken into account;
\textbf{(\textit{ii})} W/o-crf-transition and \textbf{(\textit{iii})} Small-crf-transition\footnote{Note that w.r.t. variants iii and iv, we investigate the performance of the variants under more ratios in the Appendix~\ref{Performance of More Variants}.\label{web}}: 
We ablate/deduce the CRF transition matrix. 
W/o-crf-transition denotes we do not consider the CRF transition matrix at all during training and prediction/inference; 
Small-crf-transition denotes we proceed normally during training as usual, but use $0.5$ times the value of the CRF transition matrix during prediction/inference;
\textbf{(\textit{iv})} Small-emission\footref{web}: 
We deduce the emission values, where we also proceed normally during training, but use 0.5 times the value of the emission values (e.g., the BERT’s outputs) during prediction/inference; \footnote{Note that it is not feasible to completely ablate emission values in the prediction, because we need to take sentence sequence $\mathbf{x}$ to predict truth sequence $\mathbf{t}$.} 
\textbf{(\textit{v})} Other-clasifier-init: 
We perform the possibly inadequate learning of the parameters of the classifier part during initialization in an attempt to obtain a weaker initialization ($50$ back-propagations on the CoNLL-03 (MTurk) dataset and one epoch learning on the other three datasets);
\textbf{(\textit{vi})} Other-worker-init: We initialize the diagonal/non-diagonal elements of the weak source transition matrix to $1/classes$/$0$, respectively;
\textbf{(\textit{vii})} Other-both-init: 
We use both of the parameter initialization ways above;
\textbf{(\textit{viii})} Freeze-source:
We freeze the learning of the weak source parameters in the training phase.

In Table~\ref{Table:7}, we see that:
\textbf{(\textit{i})}
The method shows substantial performance degradation when either the weak source transition matrix or CRF transition matrix are ablated, or emission values are attenuated; 
\textit{these results directly indicate the indispensable role of all three modules}
(i.e., the weak source transition matrix, the CRF transition matrix, and the emission value) and the most significant of the weak source transition matrix;
\textbf{(\textit{ii})}
Further, smaller emission values relative to a smaller CRF transition matrix produce a more pronounced performance degradation, illustrating the more dramatic sensitivity for emission values of our method;
\textbf{(\textit{iii})}
On most of the datasets, our initialization of the classifier part and the weak source part is effective; a suitable parameter initialization allows our model to achieve better performance;
\textbf{(\textit{iv})}
Further learning of the weak source parameters is necessary for the learning process of Neural-Hidden-CRF;
\textbf{(\textit{v})}
In addition, we find that the vairant Other-classifier-init outperforms Neural-Hidden-CRF in prediction on CoNLL-03 (MTurk).
This is not surprising because we do not perform detailed tuning of our method, and the seemingly weaker parameter initialization happens to have stronger performance when combined with other hyperparameters.

\section{Conclusion}
\label{conclusion}
This paper presents Neural-Hidden-CRF, the first neuralized undirected graphical model, for learning from weak-supervised sequence labels.
Neural-Hidden-CRF embedded with a hidden CRF layer models the variables of word sequence, latent ground truth sequence, and weak label sequence, where truth sequence is provided with rich contextual semantic information by the deep learning model.
Our method, therefore, benefits both from the principled modeling of graphical models and from contextual knowledge of deep learning models, while avoiding the label bias problem caused by the local optimization perspective.
Our empirical evaluations on multiple benchmarks demonstrate that Neural-Hidden-CRF significantly improves  state-of-the-art and  provides a new solution to weakly-supervised sequence labeling.

\begin{acks}
This work was supported by National Natural Science Foundation of China Under Grant Nos (61972013, 61932007, 62141209).
\end{acks}

\bibliographystyle{ACM-Reference-Format}
\balance
\bibliography{ref}

\appendix
\onecolumn
\setcounter{figure}{0}
\setcounter{table}{0}
\renewcommand{\thefigure}{A.\arabic{figure}}
\renewcommand{\thetable}{A.\arabic{table}}
\section{Appendix}
\label{Appendix}

\subsection{Calculation of the Likelihood}
\label{Calculation of the Likelihood}
First, we have the likelihood:
\begin{small}
\begin{equation}
\begin{aligned}
& \log p(\mathbf{y}^{(i)} \mid \mathbf{x}^{(i)}; \Theta) \\
=& \log \sum_{\mathbf{t}^{(i)}}   p(\mathbf{y}^{(i)}, \mathbf{t}^{(i)} \mid \mathbf{x}^{(i)}; \Theta) \\
=&\log  \frac{1}{\boldsymbol{Z}(
\mathbf{x}^{(i)}; \Theta)} \sum_{\mathbf{t}^{(i)}}  \exp \left(\sum_{l}\sum_{w} \theta_{w} \cdot f_{w}(\mathbf{y}^{(i)}_{l},
t^{(i)}_{l-1},
t^{(i)}_{l},
\mathbf{x}^{(i)}, l)\right)\\
=&\log  \sum_{\mathbf{t}^{(i)}}  \exp \left(\sum_{l}\sum_{w} \theta_{w} \cdot f_{w}(\mathbf{y}^{(i)}_{l},
t^{(i)}_{l-1},
t^{(i)}_{l},\mathbf{x}^{(i)}, l)\right)
- \log \sum_{\mathbf{y}^{(i)}} \sum_{\mathbf{t}^{(i)}} \exp \left(\sum_{l}\sum_{w} \theta_{w} \cdot f_{w}(\mathbf{y}^{(i)}_{l},
t^{(i)}_{l-1},
t^{(i)}_{l},
\mathbf{x}^{(i)}, l)\right)\\
=&\log  \sum_{\mathbf{t}^{(i)}}  \exp (\Psi(\mathbf{y}^{(i)}, \mathbf{t}^{(i)}, \mathbf{x}^{(i)}))
- \log  \sum_{\mathbf{y}^{(i)}} \sum_{\mathbf{t}^{(i)}}  \exp (\Psi(\mathbf{y}^{(i)}, \mathbf{t}^{(i)}, \mathbf{x}^{(i)})),
\end{aligned}
\label{A:1}
\tag{A.1}
\end{equation}
\end{small}
where we use $\Psi(\mathbf{y}^{(i)}, \mathbf{t}^{(i)}, \mathbf{x}^{(i)})$ to implement the abbreviation.
We present the detail calculations of $\log \sum_{\mathbf{t}^{(i)}} \exp (\Psi(\mathbf{y}^{(i)}, \mathbf{t}^{(i)}, \mathbf{x}^{(i)}))$ and $\log \sum_{\mathbf{y}^{(i)}} \sum_{\mathbf{t}^{(i)}} \exp (\Psi(\mathbf{y}^{(i)}, \mathbf{t}^{(i)}, \mathbf{x}^{(i)}))$ in the following.

\subsubsection{Calculation of $\log \sum_{\mathbf{t}^{(i)}} \exp (\Psi(\mathbf{y}^{(i)}, \mathbf{t}^{(i)}, \mathbf{x}^{(i)}))$}

First, we define
$\boldsymbol{\alpha}^{(i)}_{l, k} \triangleq \log \left[
\sum_{\mathbf{t}^{(i)}_{1 \sim l-1}}
\exp (\Psi(\mathbf{y}^{(i)}_{1 \sim l},
\mathbf{t}^{(i)}_{1 \sim l-1},
t^{(i)}_l=k,
\mathbf{x}^{(i)}_{1 \sim l}
))\right]$,
which is used to express the logarithm of the cumulative sum of the scores after the exponential operation for each path that satisfies ``the state of  $\mathbf{t}^{(i)}$ at time step $l$ is $k$''; here the path is considered only from the beginning to the time step $l$.

Then, we have:
\begin{small}
\begin{equation}
\begin{aligned}
& \log \left[\sum_{\mathbf{t}^{(i)}} \exp (\Psi(\mathbf{y}^{(i)}, \mathbf{t}^{(i)}, \mathbf{x}^{(i)}))\right] \\
=& \log \left[
\sum_{\mathbf{t}^{(i)}_{1 \sim L}} \exp (\Psi(\mathbf{y}^{(i)}_{1 \sim L}, \mathbf{t}^{(i)}_{1 \sim L},
\mathbf{x}^{(i)}_{1 \sim L}))\right] \\
=& \log \left[\sum_{k=1}^K
\sum_{\mathbf{t}^{(i)}_{1 \sim L-1}} \exp (\Psi(\mathbf{y}^{(i)}_{1 \sim L}, \mathbf{t}^{(i)}_{1 \sim L},
t_{L}=k,
\mathbf{x}^{(i)}_{1 \sim L}))\right] 
\\
=& \log \left[\sum_{k=1}^K \exp (\boldsymbol{\alpha}^{(i)}_{L-1, k})\right]. 
\end{aligned}
\label{A:2}
\tag{A.2}
\end{equation}
\end{small}
Thus, we transform the original objective of calculating $\log \sum_{\mathbf{t}^{(i)}} \exp (\Psi(\mathbf{y}^{(i)}, \mathbf{t}^{(i)}, \mathbf{x}^{(i)}))$ into the calculating the
``log\_sum\_exp'' (i.e., the successive operations of $\exp (\cdot)$, cumulative calculation and $\log (\cdot)$) of vector $\boldsymbol{\alpha}^{(i)}_{L-1, :}$.

Now we use dynamic programming to calculate $\boldsymbol{\alpha}^{(i)}$.
The recursive calculation of $\boldsymbol{\alpha}^{(i)}$ is as follows:
\begin{small}
\begin{equation}
\begin{aligned}
\boldsymbol{\alpha}^{(i)}_{l, k} \triangleq
& 
 \log \left[\sum_{\mathbf{t}^{(i)}_{1 \sim l-1}} \exp (\Psi(\mathbf{y}^{(i)}_{1 \sim l}, \mathbf{t}^{(i)}_{1 \sim l-1}, t_l=k, \mathbf{x}^{(i)}_{1 \sim l}))\right]
\\
=& 
\log \left[\sum_{k^{\prime}=1}^K
\sum_{\mathbf{t}_{1 \sim l-2}} \exp \left(\Psi(\mathbf{y}^{(i)}_{1 \sim l-1}, \mathbf{t}^{(i)}_{1 \sim l-2},
t^{(i)}_{l-1}=k^{\prime},
\mathbf{x}^{(i)}_{1 \sim l-1})
+ E^{(i)}_{k, l}
+ T_{k^{\prime}, k}
+ W_{k, \mathbf{y}^{(i)}_l}
\right)\right] 
\\
=& 
\log \left[\sum_{k^{\prime}=1}^K
 \exp (\boldsymbol{\alpha}^{(i)}_{l-1, k^{\prime}}
+ E^{(i)}_{k, l}
+ T_{k^{\prime}, k}
+ W_{k, \mathbf{y}^{(i)}_l}
)\right], 
\end{aligned}
\label{A:3}
\tag{A.3}
\end{equation}
\end{small}
where $W_{k, \mathbf{y}_l^{(i)}}=\sum_{j \in \mathcal{J}^{(i)}} \pi_{k, y_l^{(i, j)}}^{(j)}$, and $E^{(i)}_{k, l}$, $T_{k^{\prime}, k}$, $W_{k, \mathbf{y}^{(i)}_l}$ denote the
\textit{emission score}, the \textit{CRF transition score}, and the \textit{weak source transition score}, which originate from the three kind of feature functions.

The boundary case of $\boldsymbol{\alpha}^{(i)}$ is:
\begin{small}
\begin{equation}
\boldsymbol{\alpha}^{(i)}_{0, k}
=\left\{\begin{array}{ll}
\underbrace{0}_{\text{obtained from $\log(1)$}}
& \text { if } 
k=
\text {BEGIN} \\
\underbrace{-10000}_{\text{replace $\log(0)=-\infty$}} & \text { otherwise}.
\end{array}\right.
\label{A:4}
\tag{A.4}
\end{equation}
\end{small}

\subsubsection{Calculation of 
$\log \sum_{\mathbf{y}^{(i)}} \sum_{\mathbf{t}^{(i)}} \exp (\Psi(\mathbf{y}^{(i)}, \mathbf{t}^{(i)}, \mathbf{x}^{(i)}))$}

First, we define
$\boldsymbol{\beta}^{(i)}_{l, k} \triangleq \log \left[
\sum_{\mathbf{y}^{(i)}_{1 \sim l}}
\sum_{\mathbf{t}^{(i)}_{1 \sim l-1}}
\exp (\Psi(\mathbf{y}^{(i)}_{1 \sim l},
\mathbf{t}^{(i)}_{1 \sim l-1},
t^{(i)}_l=k,
\mathbf{x}^{(i)}_{1 \sim l}
))\right]$.
Similar to the derivation in Equation~\ref{A:2}, we can do the following derivation for $\log \sum_{\mathbf{y}^{(i)}} \sum_{\mathbf{t}^{(i)}} \exp (\Psi(\mathbf{y}^{(i)}, \mathbf{t}^{(i)}, \mathbf{x}^{(i)}))$.

Thus, we have:
\begin{small}
\begin{equation}
\begin{aligned}
& \log \left[ \sum_{\mathbf{y}^{(i)}}\sum_{\mathbf{t}^{(i)}} \exp (\Psi(\mathbf{y}^{(i)}, \mathbf{t}^{(i)}, \mathbf{x}^{(i)}))\right] \\
=& \log \left[
\sum_{\mathbf{y}^{(i)}_{1 \sim L}}
\sum_{\mathbf{t}^{(i)}_{1 \sim L}} \exp (\Psi(\mathbf{y}^{(i)}_{1 \sim L}, \mathbf{t}^{(i)}_{1 \sim L},
\mathbf{x}^{(i)}_{1 \sim L}))\right] \\
=& \log \left[\sum_{k=1}^K
\sum_{\mathbf{y}^{(i)}_{1 \sim L}}
\sum_{\mathbf{t}^{(i)}_{1 \sim L-1}} \exp (\Psi(\mathbf{y}^{(i)}_{1 \sim L}, \mathbf{t}^{(i)}_{1 \sim L},
t^{(i)}_{L}=k,
\mathbf{x}^{(i)}_{1 \sim L}))\right] 
\\
=& \log \left[\sum_{k=1}^K \exp (\boldsymbol{\beta}^{(i)}_{L-1, k})\right]. 
\end{aligned}
\label{A:5}
\tag{A.5}
\end{equation}
\end{small}
The recursive calculation of $\boldsymbol{\beta}^{(i)}$ is:
\begin{small}
\begin{equation}
\begin{aligned}
\boldsymbol{\beta}^{(i)}_{l, k} \triangleq
& 
 \log \left[
 \sum_{\mathbf{y}^{(i)}_{1 \sim l}}
 \sum_{\mathbf{t}^{(i)}_{1 \sim l-1}} \exp (\Psi(\mathbf{y}^{(i)}_{1 \sim l}, \mathbf{t}^{(i)}_{1 \sim l-1}, t^{(i)}_l=k, \mathbf{x}^{(i)}_{1 \sim l}))\right]
\\
=& 
\log \left[\sum_{k^{\prime}=1}^K
 \sum_{\mathbf{y}^{(i)}_{1 \sim l-1}}
\sum_{\mathbf{t}^{(i)}_{1 \sim l-2}} 
\sum_{\mathbf{y}^{(i)}_{l}}
\exp \left(\Psi(\mathbf{y}^{(i)}_{1 \sim l-1}, \mathbf{t}^{(i)}_{1 \sim l-2},
t^{(i)}_{l-1}=k^{\prime},
\mathbf{x}^{(i)}_{1 \sim l-1})
+ E^{(i)}_{k, l}
+ T_{k^{\prime}, k}
+ W_{k, \mathbf{y}^{(i)}_l}
\right)\right] 
\\
=& 
\log \left[\sum_{k^{\prime}=1}^K
 \sum_{\mathbf{y}^{(i)}_{1 \sim l-1}}
\sum_{\mathbf{t}^{(i)}_{1 \sim l-2}} 
\exp \left(\Psi(\mathbf{y}^{(i)}_{1 \sim l-1}, \mathbf{t}^{(i)}_{1 \sim l-2},
t^{(i)}_{l-1}=k^{\prime},
\mathbf{x}^{(i)}_{1 \sim l-1})
+ E^{(i)}_{k, l}
+ T_{k^{\prime}, k}
+ 
\log \sum_{\mathbf{y}^{(i)}_l} \exp (W_{k, \mathbf{y}^{(i)}_l})
\right)\right] 
\\
=& 
\log \left[\sum_{k^{\prime}=1}^K
 \exp \left(\boldsymbol{\beta}^{(i)}_{l-1, k^{\prime}}
+ E^{(i)}_{k, l}
+ T_{k^{\prime}, k}
+ 
\log \sum_{\mathbf{y}^{(i)}_l} \exp (W_{k, \mathbf{y}^{(i)}_l})
\right)\right],
\end{aligned}
\label{A:6}
\tag{A.6}
\end{equation}
\end{small}
where the meanings of  $E^{(i)}_{k, l}$, $T_{k^{\prime}, k}$, $W_{k, \mathbf{y}^{(i)}_l}$ are the same as those of the corresponding symbols in Equation~\ref{A:3}.
Also, the boundary case is:
\begin{small}
\begin{equation}
\boldsymbol{\beta}^{(i)}_{0, k}
=\left\{\begin{array}{ll}
\underbrace{0}_{\text{obtained from $\log(1)$}}
& \text { if } 
k=
\text {BEGIN} \\
\underbrace{-10000}_{\text{replace $\log(0)=-\infty$}} & \text { otherwise}.
\end{array}\right.
\label{A:7}
\tag{A.7}
\end{equation}
\end{small}
Specifically, in Equation~\ref{A:6}, we can also use dynamic programming to  calculate $\log \sum_{\mathbf{y}^{(i)}_l} \exp (W_{k, \mathbf{y}^{(i)}_l})$.

\subsection{Probabilistic Graphical Representation}
\label{Probabilistic Graphical Representation}

\begin{figure*}[ht]
\centering
\includegraphics[width=0.95\textwidth]{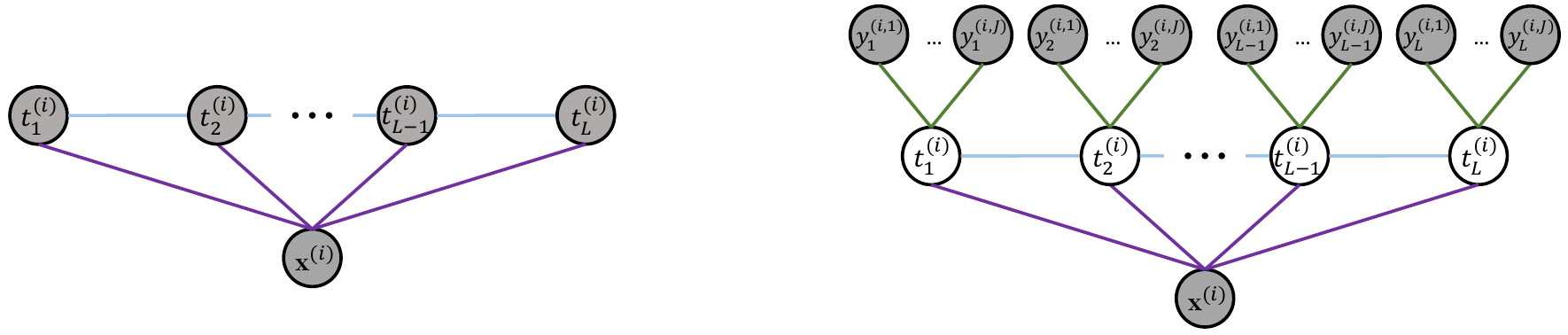} 
\caption{Probabilistic graphical representation of CRF (left) vs. Neural-Hidden-CRF (right).}
\end{figure*}

\subsection{Experimental Configurations}
\label{Experimental Configurations}

\begin{table*}[h]
\begin{minipage}{0.5\linewidth}
\fontsize{9pt}{9pt}\selectfont
\begin{threeparttable}[b]
\caption{Configurations  on CoNLL-03 (MTurk).}
\begin{tabular}{  l l l l l}
\toprule
&  \texttt{Batch}$^{\S}$
  & \texttt{Lr}$^{*}$  
   & \texttt{Lr\_weak}$^{\dag}$  & \texttt{\(\rho\)}$^{\dag}$
 \tabularnewline
\midrule
MV + BiLSTM-CRF
& 
32 & 0.1  & - & -
\\
MV + BiLSTM & 1  &  0.01  & - & - \\
CL (VW)
& 32  & 0.1  & - & - \\
CL (VW-B)
  &  128   &0.001  & - & - \\
CL (MW)
 & 32  &0.1  & - & - \\
Neural-Hidden-CRF
 & 64 & 0.1   & 0.0001 & 2.0 \\
Gold + BiLSTM-CRF
 & 1   & 0.01 & - & -
\tabularnewline
\bottomrule
\end{tabular}
\begin{tablenotes}
\item[1] $\S$: 
The search space of batch size is: (1, 32, 64, 128).
\item[2] $*$: 
The search space of learning rate is: (0.1, 0.01, 0.001).
\item[3] $\dag$: 
The search spaces for our specific hyper-parameters are:
Lr\_weak: (0.001, 0.0001),  $\rho$: (2.0).
\end{tablenotes}
\end{threeparttable}
\end{minipage}
\begin{minipage}{0.48\linewidth}
\centering
\fontsize{9pt}{9pt}\selectfont
\begin{threeparttable}
\caption{Configurations of  Neural-Hidden-CRF on CoNLL-03 (WS), WikiGold (WS) and MIT-Restaurant (WS).}
\begin{tabular}{  l l l l l l}
\toprule
&  \texttt{Batch}$^{\S}$
  & \texttt{Lr}$^{\S}$
   & \texttt{Lr\_crf}$^{\S}$ 
   & \texttt{Lr\_weak}$^{*}$
   & \texttt{\(\rho\)}$^{*}$
 \tabularnewline
\midrule
Conll (P)
& 32 &  2e-5  & 0.001 & 0.001 & 2.0
\\
Conll (I)& 32  & 2e-5   &0.01  &0.001  &2.0  \\
Wikigold (P)
& 16 & 2e-5  & 0.005 & 0.001 & 2.0\\
Wikigold (I)
  &  32 & 3e-5 & 0.001  &0.001  &2.0 \\
Mit-Rest. (P)
 & 32  & 2e-5  & 0.001  &0.01  &6.0  \\
Mit-Rest. (I)
 & 16  &2e-5  &0.01  &0.2  &5.0
\tabularnewline
\bottomrule
\end{tabular}
\begin{tablenotes}
\item[1] $\S$: 
The search spaces  are consistent with benchmark~\cite{zhang2021wrench}.
\item[2] $*$:
The search spaces  for our specific hyper-parameters on Mit-Rest are:
Lr\_weak=(0.2,0.1,0.01), $\rho$=(5.0, 6.0).
On other datasets, our model more easily yields satisfactory results, and we always set them (0.001, 2.0).
\end{tablenotes}
\end{threeparttable}
\captionsetup{type=table} 
\end{minipage}
\end{table*}

\subsection{Suggestions for Setting Hyperparameters}
\label{Suggestions for Setting Hyperparameters}

When applying our Neural-Hidden-CRF to other datasets, in most cases, we recommend considering the following suggestions for setting hyperparameters.

\begin{itemize}
\item
For \texttt{Batch} (batch size):
Our suggested finding space is $\{8, 16, 32, 64, ...\}$, and batch size should not be set to $1$ (which would not be conducive to the challenging multi-source weak supervision learning context);
\item 
For \texttt{Lr\_weak} (learning rate of weak source transition matrix):
We suggest that \texttt{Lr\_weak} be set equal to or less than the learning rate of the CRF layer (i.e., \texttt{Lr\_crf});
\item 
For \texttt{\(\rho\)} (in Equation~\ref{equation 21}):
Our suggested finding space is $\{2.0, 3.0, 4.0, 5.0, 6.0\}$ for most cases;
\item 
For the pre-train of the classifier part of the model (mentioned in Section~\ref{Implementation Details}):
We suggest using better super-parameters (e.g., batch size, learning rates, etc.) for pre-training to get a better parameter initialization.
\end{itemize}

\subsection{Performance of More Variants}
\label{Performance of More Variants}

\begin{figure*}[ht]
\centering\includegraphics[width=0.95\textwidth]{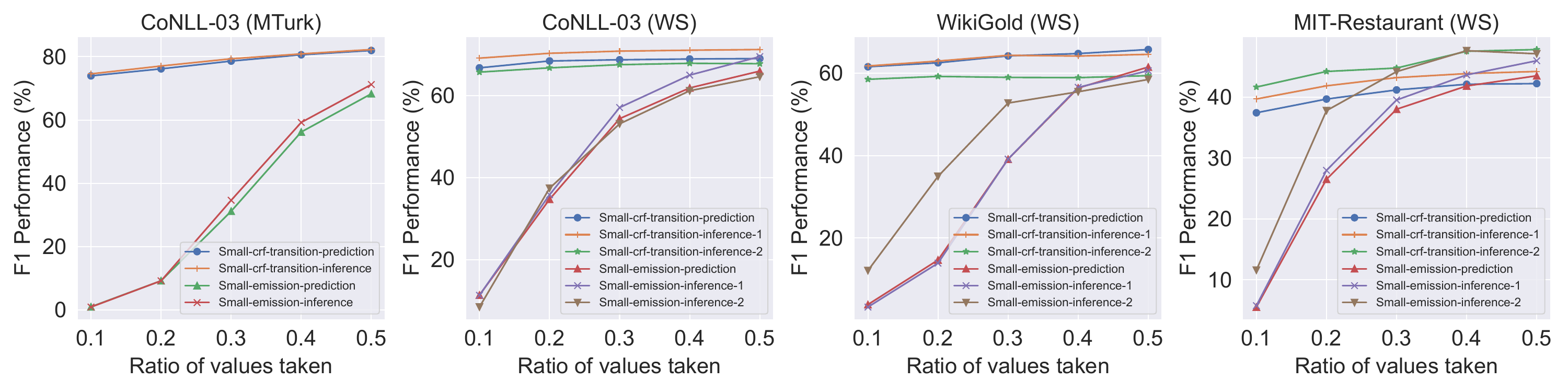} 
\caption{Performance of more variants for supplementary ablation study. Results are averaged over 20 runs.}
\label{Figure:A1}
\end{figure*}

\end{document}